\documentclass[10pt,twocolumn,letterpaper]{article}

\usepackage{cvpr}      % To produce the REVIEW version
\definecolor{cvprblue}{rgb}{0.21,0.49,0.74}
\usepackage[pagebackref,breaklinks,colorlinks,allcolors=cvprblue]{hyperref}
\usepackage{chngcntr}
\usepackage{graphicx}
\usepackage{multirow}
\usepackage{xcolor}
\usepackage{colortbl}
\usepackage{makecell} % To create multi-line cells
\usepackage{amssymb}
\usepackage{bbding}
\usepackage{pifont}
\usepackage{wasysym}
\usepackage{utfsym}
\usepackage{fontawesome}
\usepackage[pagebackref,breaklinks,colorlinks]{hyperref}
\usepackage[capitalize]{cleveref}
\definecolor{my_green}{RGB}{51,102,0}
\definecolor{my_red}{RGB}{204, 0, 0}
\renewcommand{\checkmark}{\textcolor{my_green}{\ding{51}}} % ✔
\newcommand{\crossmark}{\textcolor{my_red}{\ding{55}}} % ✘

\definecolor{ModelGreen}{RGB}{213,232,212}

\def\paperID{8650} % *** Enter the Paper ID here
\def\confName{CVPR}
\def\confYear{2025}

\author{
Yan Shu\textsuperscript{1,2} \quad
Zheng Liu\textsuperscript{2\thanks{Correspondence to \texttt{<bo.zhao@sjtu.edu.cn>} and \texttt{<zhengliu1026@gmail.com>}}} \quad
Peitian Zhang\textsuperscript{3} \quad 
Minghao Qin\textsuperscript{4} \quad \\
Junjie Zhou\textsuperscript{5} \quad
Zhengyang Liang\textsuperscript{2} \quad
Tiejun Huang\textsuperscript{6} \quad
Bo Zhao\textsuperscript{1,2*} \quad \\
[2mm]
\textsuperscript{1}~Shanghai Jiaotong University \quad \textsuperscript{2}~Beijing Academy of Artificial Intelligence \quad \textsuperscript{3}~Renmin University of China\\
\textsuperscript{4}~Chinese Academy of Sciences \quad
\textsuperscript{5}~Beijing University of Posts and Telecommunications 
\textsuperscript{6}~Peking University
\\
[2mm]
\normalsize{
\url{https://github.com/VectorSpaceLab/Video-XL}
}
}

\usepackage{amssymb}
\title{\includegraphics[width=0.8cm]{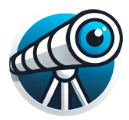}Video-XL: Extra-Long Vision Language Model for
Hour-Scale Video Understanding}

\newcommand{\vst}{\langle\text{vs}\rangle}

\begin{document}

\maketitle

\begin{abstract}

Long video understanding poses a significant challenge for current Multi-modal Large Language Models (MLLMs). Notably, the MLLMs are constrained by their limited context lengths and the substantial costs while processing long videos. Although several existing methods attempt to reduce visual tokens, their strategies encounter severe bottleneck, restricting MLLMs' ability to perceive fine-grained visual details. 
In this work, we propose \textbf{Video-XL}, a novel approach that leverages MLLMs' inherent key-value (KV) sparsification capacity to condense the visual input. Specifically, we introduce a new special token, the Visual Summarization Token (\textbf{VST}), for each interval of the video, which summarizes the visual information within the interval as its associated KV. The VST module is trained by instruction fine-tuning, where two optimizing strategies are offered. 1. \textbf{Curriculum learning}, where VST learns to make small (easy) and large compression (hard) progressively. 2. \textbf{Composite data curation}, which integrates single-image, multi-image, and synthetic data to overcome the scarcity of long-video instruction data. The compression quality is further improved by \textbf{dynamic compression}, which customizes compression granularity based on the information density of different video intervals. 
Video-XL's effectiveness is verified from three aspects. First, it achieves a superior long-video understanding capability, outperforming state-of-the-art models of comparable sizes across multiple popular benchmarks. Second, it effectively preserves video information, with minimal compression loss even at 16$\times$ compression ratio. Third, it realizes outstanding cost-effectiveness, enabling high-quality processing of thousands of frames on a single A100 GPU. 

% Our code is available in this anonymous repo. 
\end{abstract}

\section{Introduction}
\label{sec:intro}

\begin{figure}[h]
    \centering
    \includegraphics[width=0.45\textwidth]{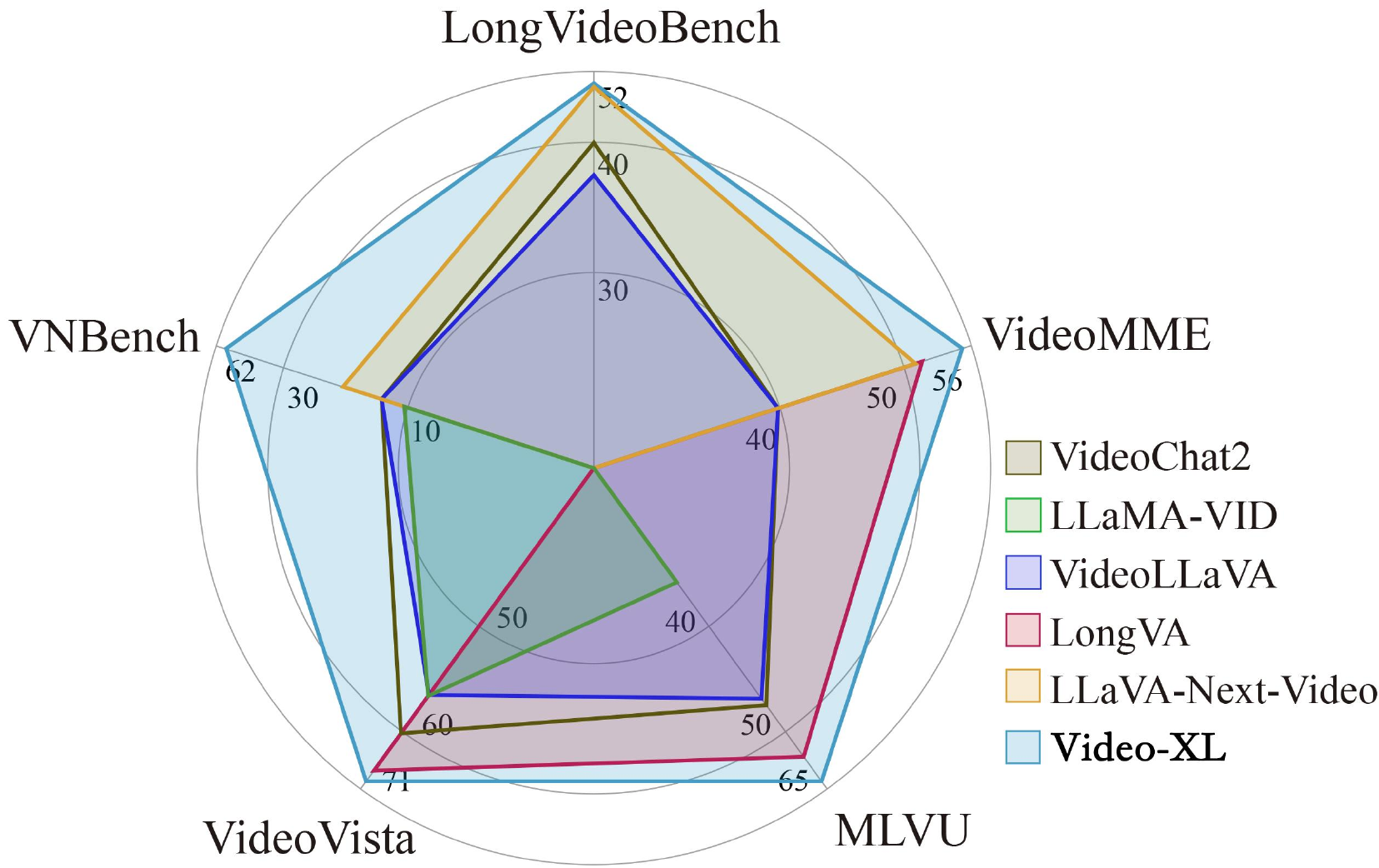}
    \caption{Video-XL achieves state-of-the-art results across several video benchmarks, surpassing other models of comparable sizes.}
    \vspace{-5pt}
    \label{fig:radar}
\end{figure}

Multi-modal Large Language Models (MLLMs) have attracted widespread attention from the AI community. By augmenting large language models (LLMs)~\cite{openai2023gpt4, ChatGPT,touvron2023llama} with vision encoders, MLLMs are enabled to perform various vision-language modeling tasks, e.g., image captioning and visual question answering~\cite{liu2023visual_llava,zhu2023minigpt4,li2023blip2}. Recently, there has been growing interest in applying MLLMs for video understanding given MLLMs' proficiency in comprehending and reasoning over visual information \cite{videollama,li2023videochat,llamavid,ye2024voco}. 

% However, constructing video-based large language models (VideoLLMs) is challenging due to the complex and detailed nature of videos, which necessitates spatial-temporal joint modeling.

% 【一个直接的方式就是将视频切分成多图，将mllms的多图能力泛化到视频域上。然而他们仅仅能处理短视频不能处理长视频，这主要是长视频会产生大量视觉token。一方面超过了语言模型窗口长度，另一方面LLM计算复杂度超过了显存限制。（举例子）】【最近一些工作尝试解决这个问题。他们采用预先压缩的方式，设计模块在LLM之前对视觉token的数量进行压缩；然而，由于压缩带来的损失，他们无法对长序列中的细节信息进行有效的信息保留。（展开：subaction, 关键情节。。）】【跟他们不同，LOngva不做压缩扩展了语言模型上下文，然而他依旧面临显存消耗的问题。（展开）】

% Technical Challenges 1.VLM short (cost and data); 2. concurrent LVU -> info bottleneck 

However, it remains a significant challenge for the existing MLLMs to process long videos due to their limited context lengths and the huge costs involved. Particularly, a long video consists of a long sequence of frames, where each frame usually consumes a large number of visual tokens for MLLMs to perceive (e.g., 144 tokens per frame). As a result, the input may easily exceed the limits of MLLMs' context lengths. Even if the context lengths can be extended sufficiently, it will still take considerable computation and memory costs to process long videos, making it challenging in real-world scenarios. Recently, many studies have attempted to reduce the token count from the visual encoder for each frame~\cite{maaz2023videochatgpt,llamavid,weng2024longvlm,moviechat2023,malmm2024}. While these approaches enable handling longer input, they often lead to a substantial loss of visual information, which creates a severe bottleneck for MLLMs' fine-grained perception of long videos.

To address the existing challenges, we introduce a novel approach for long video understanding, called \textbf{Video-XL}. Unlike the existing methods which rely on the reduction of tokens generated from visual encoder, we leverage the LLMs' inherent KV sparsification capability to generate compact representations for long videos. Specifically, it's observed that LLMs tend to form sparse attention patterns when dealing with long inputs \cite{liu2022dynamic,liu2023deja}. This phenomenon suggests that the LLMs' inputs are secretly compressed, which allows them to perceive useful information from long contexts. Based on this inspiration, we design the following compression mechanism.  

$\bullet$ \textbf{VST Compression}. We employ a new special token \textbf{VST} (\textbf{V}isual \textbf{S}ummarization \textbf{T}oken) to generate compact representations for long videos. The VST tokens are assigned to different intervals of the video, which summarizes the visual information within the intervals (i.e., the original KVs from its preceding visual tokens) into their associated KVs. The VSTs' compressed KVs are maintained for future encoding, while the KVs from other visual tokens are off-loaded. Thus, it enables a substantial cost reduction for the processing of the entire video. Knowing that different parts of a video exhibit variant information density, we propose \textbf{dynamic compression strategy}. Particularly, we form small intervals for the information-dense parts of the video; therefore, it enables fine-grained compression for the corresponding parts. On the contrary, we form large intervals for those information-sparse parts of the video, which can do with coarse-grained compression. With this operation, the visual information loss can be minimized given a fixed budget. % overall compression ratio. 

$\bullet$ \textbf{Training}. The VST module is trained by instruction fine-tuning. Given a video understanding task, the MLLM is required to generate VST compressed KVs for an input video; then, it is asked to predict the ground-truth answer based on the compression result. The training of VST is non-trivial given the challenges on the problem's complexity and the limitation of data. Therefore, we introduce the following strategies to enable the effective model training. 

First, we propose to train VST by \textbf{curriculum learning}. Once training process is started, we perform a random sampling of small compression ratios for VST (e.g., $2\times$, $4\times$). With training process going on, we gradually sample larger compression ratios for VST (e.g., $8\times$, $16\times$). As it's easier to perform small compressions, the VST module may well establish its preliminary capability after the initial stage. Upon this foundation, the VST module can progressively learn larger compressions with higher proficiency. 

Second, we employ a \textbf{composite data curation} method to create training data. Currently, long-video instruction data is still scarce; therefore, we exploit two extra resources to overcome this shortage. Considering that video understanding is built on top of the comprehension of images, we leverage \textit{single-image} and \textit{multi-image} with captioning and QA datasets for augmentation. The images are formatted as sequences of frames, which is made consistent with the video instruction data, and thus facilitates knowledge transfer. In addition, video understanding calls for precise and comprehensive retrieval of useful information for the given instruction. As a result, we create a \textit{synthetic dataset}, called VIsual Clue Ordering (VICO), to strengthen this fundamental capability. 

We implement Video-XL based on Qwen-2-7B, whose effectiveness can be verified from three perspectives. First, Video-XL outperforms state-of-the-art models of comparable sizes across popular long-video benchmarks, including VideoMME~\cite{videomme}, MLVU~\cite{zhou2024mlvu}, LongVideoBench~\cite{wu2024longvideobench}, etc., as shown in Figure~\ref{fig:radar}. Second, Video-XL realizes high-fidelity compression of long videos, as it well maintains its performance throughout various compression ratios ($2\times$, $4\times$, $8\times$, $16\times$). Third, it also produces outstanding cost-effectiveness. Notably, it effectively handles 2048 frames with a single A100 GPU, while achieving 95\% accuracy in the Needle-in-a-Haystack~\cite{longva} evaluation.

\section{Related Work}
% Building on the success of LLMs, Multimodal Large Language Models (MLLMs) introduce a visual encoder to extract multimodal features. A connector is then used to align these multimodal features to the same dimension as LLM tokens, enabling the LLM to handle multiple types of modality information. Recent advancements in MLLMs have significantly improved performance in image understanding tasks \cite{li2022blip,li2023blip2,dai2023instructblip,liu2023visual_llava,zhu2023minigpt4}. In addition, more and more work aims to extend MLLMs to a wider range of modalities. 
%% MLLM + long video understanding 去掉subsection 用 2 x paragraph，每个 1/3 单栏长度

\begin{figure*}[!t]
    \centering
    % \vspace{-0.3cm}
    \includegraphics[width=0.9\textwidth]{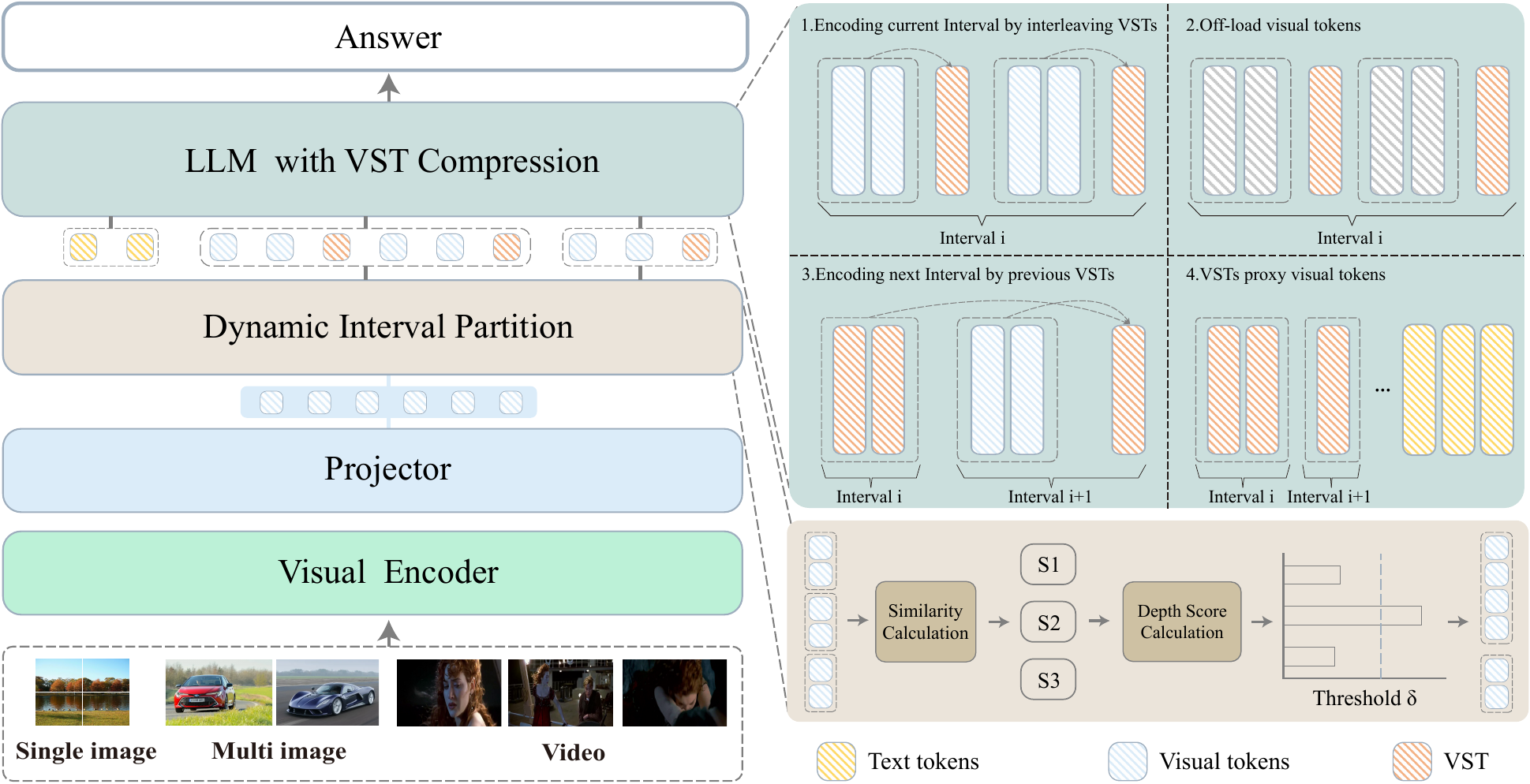}
    % \vspace{-0.5cm}
    \caption{Overview of Video-XL. The input data (single-image, multi-images, or video) is encoded and projected as visual tokens. The visual tokens are dynamically split into intervals based on semantic consistency (measured by depth score). The visual information in each interval is compressed as VSTs' KVs, which enables MLLMs to perceive and understand longer inputs than its context window.}
    % \caption{Overview of Video-XL. Video-XL utilizes a unified visual encoding scheme for single images, multi images, and videos. Visual tokens are first split into intervals by Dynamic chunk adjustor. Then, in LLM compressor, the activations from the visual contexts in each chunk are condensed into Visual Summarization Tokens (VSTs). Conditioned on the VSTs from previous chunks, Video-XL performs next-token prediction auto-regressively.  }
    \label{fig:pipe}
    \vspace{-0.3cm}
\end{figure*}

\paragraph{Multimodal Large Language Models.}
Building on the success of Large Language Models (LLMs), Multimodal Large Language Models (MLLMs) incorporate a visual encoder to extract visual features. A connector is then used to align these features to the same dimension as LLM tokens, enabling the LLM to handle visual information. Recent advancements in MLLMs~\cite{li2022blip,dai2023instructblip,zhu2023minigpt4} have significantly improved performance in image understanding tasks. As pioneers, Flamingo~\cite{alayrac2022flamingo} proposes to connect pre-trained vision-only and language-only models. The lightweight querying transformer is introduced in Blip2~\cite{li2023blip2} to bridge the gap between the image encoder and LLMs. LLaVA~\cite{liu2023visual_llava} proposes the visual instruction tuning using machine-generated instruction-following data.

\vspace{-12pt}
\paragraph{Video MLLMs.}
With the excellent foundation of image MLLMs, 
many works~\cite{maaz2023videochatgpt, luo2023valley, stllm, li2023videochat, li2024mvbench, videollama, chatunivi} try to transfer the success of image understanding to the video understanding. 
However, unlike image understanding, the main difficulty with (long) video understanding is the sheer number of tokens, which often exceeds the context length of current LLMs. To handle this, MovieChat~\cite{moviechat2023} and MA-LMM~\cite{malmm2024} use memory modules with long-term memory banks for accurate long video predictions. LLaMA-VID~\cite{llamavid} reduces each frame to two tokens with context attention, while LongVLM~\cite{weng2024longvlm} and Video-CCAM~\cite{fei2024videoccam} focus on token merging and cross-attention modules for long context modeling. But these methods suffer from serious information loss, obstructing fine-grained comprehension. Unlike these methods, LWM~\cite{liu2024world} incrementally extends context using RingAttention~\cite{liu2024blockwise}, while LongVA~\cite{longva} expands LLM context length directly. Other approaches enhance training methods~\cite{longvila} or improve LLM architecture~\cite{wang2024longllava}. However, substantial computational and memory costs for processing thousands of tokens in long videos remain unresolved.

% \clearpage

\section{Method}

% reference
% - llama-vid 
% - activation beacon
% 1. overview: MLLMs for video understanding
% - framework 
% - video encoding & visual token generation
% - prompt encoding & answer generation 
% 2. vst compression
% - compression method (参考activation beacon)
% - dynamic compression (using a better diagram?)
% 3. training
% - instruction tuning: formulation & loss function
% - curriculum learning (psuedo code?)
% - data curation 
%   > how we uniformly format the image-captioning data
%   > how we create synthetic data (motivation!)

% vst Compression
% Training techniques

\subsection{Overview}
% 结构跟llava类似，由visual encder，projector和llm组成，通过visual instruction tuning 方法训练；不同的是，videoxl在训练时设计了统一的视觉编码机制以利用不同类型的多模态数据（3.2）； 同时为了压缩大量视觉token，我们修改了LLM的工作流，使其可以无损压缩视觉token（3.3）。最后我们介绍了videoxl的training pipeline(3.4)

The architecture of Video-XL inherits the minimalism design of LLaVA series, which comprises a visual encoder, a visual-language projector, and an LLM backbone. First, the
input image is encoded by a visual encoder, where we use CLIP-ViT-L~\cite{radford2021clip} to carry out this operation.  Second, the visual encoder's output embeddings are projected as visual tokens, where we leverage a two-layer MLP component with GELU activation. Third, the visual tokens, along with the text prompts, are fed into the LLM for conditioned text generation. Video-XL is featured for the introduction of VST module, which generates compressed KVs for lightweight and thus extended processing of long videos. In this section, we'll explain details about the compression mechanism (Section~\ref{sec:vst}) and its training process (Section~\ref{sec:train}).

% Specifically, we design a VST compression mechanism to generate compact representations for long videos (see Section~\ref{sec:vst}). We then introduce our training details, including curriculum learning and composite data curation, in Section~\ref{sec:train}. 

% However, we design a unified visual encoding mechanism to represent various multimodal data in a common space (see Section~\ref{sec:visual}). Furthermore, we modify the LLM workflow to effectively compress large visual tokens to achieve promising long video understanding (see Section~\ref{sec:llm}). 

% Finally, we briefly illustrate our training pipeline in Section~\ref{sec:training_pipeline}. 
\begin{figure*}[t]
    \centering
    \vspace{-0.3cm}
    \includegraphics[width=0.9\textwidth]{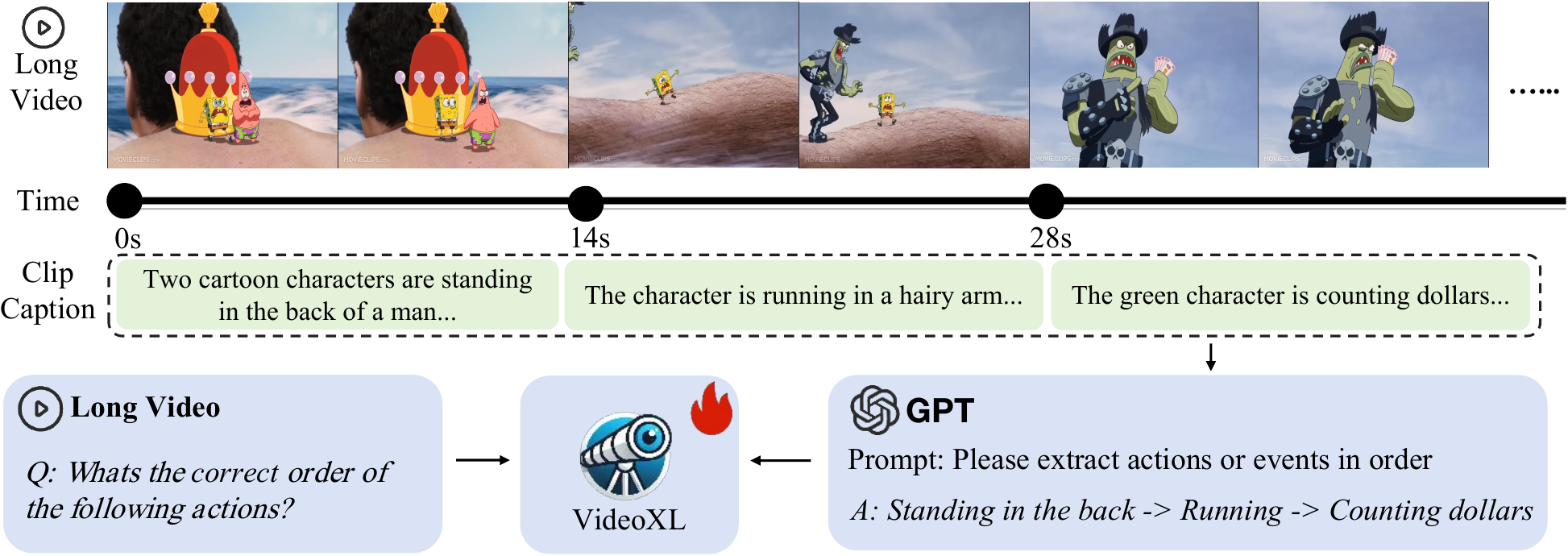}
    % \vspace{-0.5cm}
    \caption{The pipeline of VICO generation. First, we generate short-clip captions for each few-second segments of the video. Then, we use GPT to extract key actions and events following the temporal order. Finally, the long video and QA pair are presented for model training.}
    \label{fig:vico_pipe}
    \vspace{-0.3cm}
\end{figure*}

\subsection{VST Compression}\label{sec:vst}
Unlike previous methods which reduce token count before LLM, we leverage the LLM itself to generate compact representations of videos. Given visual tokens $X$, we propose to compress the KVs of $X$ into the KVs of $C$, where $|C|\ll|X|$. This could substantially save the memory cost and thus allow the model to accommodate longer visual inputs within the constraints of the LLM's context length.

\textbf{Compression mechanism}. When encoding a token $x_{i}$ within the input $X$, the LLM needs to query for the entire KVs from $X_{<i}$. Consequently, it will consume significant GPU memory due to the storage of massive visual tokens, and it will be expensive to compute due to the quadratic complexity of self-attention. To avoid the huge cost from direct computation, we partition $X$ ($\{x_1,..., x_n\}$) into shorter intervals $\{X_1, \dots, X_i\}$ of sizes $\{w_1,..., w_i\}$:
\begin{equation}
    [ x_1, \dots , x_n ] \xrightarrow{\mathrm{Partition}} [X_1, \dots, X_i],
\end{equation}
where $\sum w_i = n$  and $|X_i| = w_i$. The length of each interval is within the constraint of LLM's context window. For each interval, we introduce a new special token, called Visual Summarization Token (\textbf{VST}): $~\vst$, which prompts the LLM to compress the visual information into VST's KV, i.e. keys and values at every layer. We then determine a compression ratio $\alpha_i$ for each interval $X_i$. Based on this ratio, we uniformly interleave $k_i$ VSTs into the interval (denoted as $V_i=\{\vst^i_1,\dots,\vst^i_{k_i}\}$), where $k_i=w_i/\alpha_i$. In other words, one VST is appended to every $\alpha_i$ visual token: 
\begin{equation}
    \resizebox{0.9\columnwidth}{!}{
        $X_i\xrightarrow{\mathrm{Interleave}~V_i}X_i' = [x^i_1,\dots,x^i_{\alpha_i},\vst^i_1,~\dots~,x^i_{w_i},\vst^i_{k_i}].$
    }
\end{equation} 

The LLM encodes each of these intervals one by one. Once the encoding of $X_i$ completes, the VSTs' KVs ($V_i$) are preserved as the compression of visual information, while the visual tokens' KVs ($X_i$) are off-loaded. When encoding the next interval $X_{i+1}$, the LLM will directly condition on the accumulated KVs from all preceding VSTs ($V_{\le i}$) as a proxy to the original visual tokens $X_{\le i}$. 

% The LLM encodes these intervals \textit{one by one}. During encoding $X_i'$, Video-XL reuses all modules of the LLM except employing additional projection matrices ($\boldsymbol{W}_Q^{v}$, $\boldsymbol{W}_K^{v}$, $\boldsymbol{W}_V^{v}$) to transform VST hidden states in the self attention module of each layer. As a result, the information of the visual signal is compressed into VSTs' activations.
% After encoding $X_i'$, we \textit{discard} activations of all the raw vision tokens ($X_i$), while we \textit{accumulate} the activations of VSTs ($V_i$). When encoding the next interval $X_{i+1}'$, the LLM directly conditions on the accumulated VST activations ($V_{\le i}$) as a proxy to the raw vision tokens $X_{\le i}$.

% 在一个chunk中把视觉信息压缩到vst中是很有挑战的，因为和自然语言不同，视觉信息在时间上的分布存在不平衡。如果固定的均分chunk大小，则会在压缩过程中损失细节信息。因此我们希望利用视觉信息，将语义连贯的视觉片段放入同一个chunk中。

% 我们不使用外部模型，而是利用clip内部的知识判别视觉语义。因为vit的【cls】token代表了每帧的全局语义信息，我们计算相邻帧的相似度得到{c1,c2..ci};

%我们找到那些峰值点通过判断某帧左侧和右侧的视觉差异 

\textbf{Dynamic compression strategy.} It's trivial to divide a long video into equal-sized intervals. However, the straightforward method is suboptimal considering that the information density is variant for different parts of the video. Particularly, some parts of the video exhibit high information density (\textit{a.k.a.} information-dense), as they involve fast-changing visual semantics. While some other parts are of low information density (\textit{a.k.a.} information-sparse), as they contain slow-paced visual semantics. The information-dense portions require fine-grained compression; in contrast, the information-sparse portions can do with coarse-grained compression. Because of this property, we design a dynamic compression method which customizes the compression granularity for each interval based on its information density. Inspired by VideoLLaMB~\cite{wang2024videollamb}, we employ CLIP to estimate the changing of visual semantic. Specifically, we make use of each frame's [cls] embedding to represent its global semantic. Thus, we can calculate the similarity scores $s_i$ for two neighboring frames ($i$-th and $i$+1-th). Based on this value, we can estimate the consistency of visual semantic using the depth score (defined in \cite{wang2024videollamb}): 
\begin{equation}
    d_i=max(s_1~\dots~s_{i-1})+max(s_{i+1} ~\dots~ s_n)-2\times s_i. 
\end{equation}
Intuitively, large depth scores mean sharp changes of visual semantic, which indicates potential semantic inconsistency. In our implementation, we introduce a threshold $\delta$, where the peak scores satisfying $d_i > \delta$ are chosen as the boundaries of video intervals. This enables the information-dense parts of the video to form small intervals for fine-grained compression, while the information-sparse portions to yield big intervals for coarse-grained compression. 

% Consequently, we can leverage the peak values of $d_i$ as the boundaries of video intervals. In our implementation, we introduce a threshold $\delta$, where $d_i > \delta$

% It's challenging to distill visual information into a VST. Unlike natural language, different parts of a video exhibit variant information density, which can lead to significant information loss during compression. Therefore, we design a dynamic compression strategy which  customizes compression granularity based on the information density of different video intervals. Specifically, inspired by ~\cite{wang2024videollamb}, instead of relying on external models, we utilize the knowledge from CLIP to distinguish visual semantics, which outputs a [cls] token for each frame to represent global semantics. Thus, we can calculate the similarity scores $s_i$ for the  $i^{th}$  and  $(i+1)^{th}$ frames. Then we we identify the `peak points' by analyzing the visual disparity between the left and right sides.
% \begin{equation}
%     d_i=max(s_1~\dots~s_{i-1})+max(s_{i+1} ~\dots~ s_n)-2\times s_i
% \end{equation}

% Here, $d_i$ represents the depth score. A larger value of $d_i$ indicates a significant change between adjacent frames, suggesting a notable transition in the video content at this point. Consequently, we set a threshold $\delta$ to adaptively segment intervals that contain semantically consistent content.

\subsection{Training}
\label{sec:train}

\noindent \textbf{Objective function.} Video-XL is trained by instruction tuning, where the model learns to optimize the generation likelihood of ground-truth response conditioned on the VST's compressed KVs and the task's instruction. Formally, the generation probability of the next token is formulated as: 
\begin{equation*}
    \resizebox{\columnwidth}{!}{
        $\Pr(t_{i+1}\mid \underbrace{\vst^1_1,\dots,\vst^{j}_{k_{j}}}_{\text{compressed KVs}},
        \underbrace{s_1,\dots, s_{\text{M}}\vphantom{\vst^{\lceil n/w \rceil}_{k_{\lceil n/w \rceil}}}}_{\text{instruction}},
        \underbrace{t_1,\dots,t_{i}\vphantom{\vst^{\lceil n/w \rceil}_{k_{\lceil n/w \rceil}}}}_{\text{ground-truth}}; 
        \boldsymbol{\Theta}),$
    }
\end{equation*}
where $\boldsymbol{\Theta}$ denotes the learnable parameters of the MLLM and VST module. We perform standard auto-regression to train the model, which minimizes the prediction loss for each of the tokens in ground-truth response.  

% denotes the parameters of the LLM, the vision encoder, the projector, the matrices for VSTs at each layer, $\boldsymbol{W}_Q^{v}$, $\boldsymbol{W}_K^{v}$, $\boldsymbol{W}_V^{v}$, and the token embedding of VST, $\boldsymbol{e}_{\vst}$ (we use one shared embedding for all VSTs). The standard auto-regression loss is minimized for training. Note that we exclude the VSTs from the training loss (setting their labels to -100) because they are only intended for compression.

\textbf{Curriculum learning.} The VST module is expected to support a wide range of compression ratios so as to flexibly handle videos of different lengths. 
By comparison, it's challenging to perform substantial compressions of long videos; however, it can be much easier to make small compressions. Because of this property, we propose to train Video-XL through curriculum learning. When the training process is started, we randomly sample small compression ratios, e.g., from (2, 4). Based on the sampled ratio, we apply VST compression to the input video and train the model via instruction tuning. In this stage, the VST module can acquire a preliminary capability in summarizing the visual information, which establishes a solid foundation to handle larger compressions. After the initial stage, we gradually improve the candidate compression ratios to 8, 12, and 16, thereby extending VST's capability in making larger compression.

% Generating text conditioned on compressed visual information is challenging, particularly with high compression ratios. To address this, we propose training Video-XL using curriculum learning. Specifically, we automatically divide the training process into three equal stages based on data size. In the initial stage, we randomly sample compression ratios from the range (2, 4) for VST. As training progresses, we increase the sampled compression ratios to (8, 12). In the final stage, we expand the compression ratio to 16. On one hand, this approach allows VST to progressively learn more aggressive compressions with high proficiency. On the other hand,  one can flexibly choose one compression ratio for all intervals according to the specific efficiency requirement.

% A naive way to train the LLM compressor is to set a dedicated compression ratio. However, different chunks contain distinct visual redundancy. Thus,  we adopt a flexible strategy in which $\alpha_i$ for the $i^{th}$ chunk is randomly sampled from $\{2, 4, 8, 12, 16\}$ during training. 
% Inspired by curriculum learning, we utilize a smaller ratio from $\{2, 4, 8\}$ during the initial one-third of training, and increase it to 12 and 16 as training advances.  At inference, one can flexibly choose one compression ratio for all chunks according to the specific efficiency requirement.

\begin{table*}[t]\small
\centering

\vspace{-4mm}
\addtolength\tabcolsep{-2.4pt} 
\resizebox{1.0\linewidth}{!}{
\begin{tabular}{lc|cc|cc|cc|c|c|c|c|c}
\toprule
\multicolumn{1}{c}{\multirow{2}{*}{Model}} & \multicolumn{1}{c|}{\multirow{2}{*}{Size}} & \multicolumn{2}{c|}{MLVU Dev}  & \multicolumn{2}{c|}{MLVU Test}& \multicolumn{2}{c|}{VideoMME}& \multicolumn{1}{c|}{\multirow{2}{*}{VNBench}}  & \multicolumn{1}{c|}{\multirow{2}{*}{VideoVista}} & \multicolumn{1}{c|}{\multirow{2}{*}{LongVideo.}}& \multicolumn{1}{c|}{\multirow{2}{*}{VideoChat.}}& \multicolumn{1}{c}{\multirow{2}{*}{MVBench}}\\
\multicolumn{1}{c}{} & \multicolumn{1}{c|}{}  & M-avg & G-avg & M-avg & G-avg & W/o sub    & W sub & \multicolumn{1}{|c|}{}      & \multicolumn{1}{|c|}{}  & \multicolumn{1}{|c|}{} & \multicolumn{1}{|c|}{} & \multicolumn{1}{c}{} \\ \midrule
\rowcolor{gray!15}\multicolumn{13}{c}{\textbf{Proprietary Models}} \\
GPT-4V~\cite{openai2023gpt4}   & - & 49.2 & 5.35 &43.3 &4.67  &59.5& 63.3 & 48.9 & -   & 59.1 & \textbf{4.06} &\textbf{43.5}   \\
GPT-4o~\cite{gpt4o}  & - & \textbf{64.6} & \textbf{5.80} & \textbf{54.9}&\textbf{5.87}   &71.9& 71.2 & 64.4 &\textbf{78.3}   & \textbf{66.7} & - &- \\
Gemini-1.5-Pro~\cite{reid2024gemini} & - & - & -&- &  -  &\textbf{75.0}& \textbf{81.3} &  \textbf{66.7} &-   & 64.0  & - &- \\
% Qwen2-VL~\cite{wang2024qwen2vl} & 7B & - & -&- &  -  &63.3&69.0 & 33.9 &75.6  & 55.6 & - &\textbf{67.0}  \\
\midrule
\rowcolor{gray!15}\multicolumn{13}{c}{\textbf{Open-source MLLMs}} \\ 
VideoChat2~\cite{li2024mvbench} & 7B & 47.9 & 3.99 &35.1 &\underline{3.99}  &39.5& 43.8 &12.4 & 61.6 & 39.3 &2.98 &\underline{62.3}   \\
LLaMA-VID~\cite{llamavid} & 7B & 33.2 & 4.22   &17.2 &3.43  &-& - & 10.8   & 56.9 & - &2.89 &41.4 \\
% Chat-UniVi~\cite{chatunivi} & 7B &- & - &-   &-& 40.6 & 45.9   & - &54.2  & - &2.99 &-  \\
VideoLLaVA~\cite{videollava} & 7B & 47.3 & 3.84 &30.7 &3.68  & 39.9 & 41.6   &12.4  &56.6 &39.1 &2.84 &43.0 \\
ST-LLM~\cite{stllm} & 7B & - & - &- &-  &37.9& 42.3 & 22.7  &49.3 & -  &3.15 &54.9 \\
Shargpt4Video~\cite{chen2024sharegpt4video} & 7B & 46.4 & 3.77 &33.8 &3.63  & 39.9& 43.6 & -  &53.6 & 39.7 &- &51.2  \\
LLaVA-Next-Video~\cite{zhang2024llavanextvideo} & 34B & - & - &- &-  & 52.0& \underline{54.9} & 20.1 &56.7 & \underline{50.5} &\textbf{3.26} &-   \\
% InternVL-Chat-V1.5~\cite{internvl} & 20B & 50.4 & 4.02 & 37.3& 3.96  & 50.7& 52.4    & - &- &\textbf{51.2} &- &-   \\
PLLaVA~\cite{xu2024pllava} & 7B & - & - & - & -  & -&-    & - &60.4 &40.2 &3.12 &46.6   \\
LongVA\dag~\cite{longva} & 7B & 56.3 & \underline{4.33} &41.1 &3.91  &\underline{52.6} & 54.3 &41.5&67.4 & 47.8  &- &- \\
% VILA~\cite{lin2024vila} & 34B & 56.7 & 4.31 & \textbf{72.0} & \textbf{61.2} &\textbf{53.8} & \textbf{62.3} & \textbf{74.0} & \textbf{62.6} & \textbf{55.7} & \text64.1\\
% VITA~\cite{fu2024vita}& 8x7B & - & - &- &-  &55.8 & 59.2& - & -  &- &-  \\
VideoLLaMA2\dag~\cite{cheng2024videollama2}& 8x7B & - & - &- &-  &47.9 & 49.7& 24.9 &60.5 & 36.0  &\textbf{3.26} &53.9  \\
% LongVILA~\cite{longvila}& 7B & - & - &- &-  &61.8 & 49.7& 39.7 & 50.5  &  \\
Video-CCAM\dag~\cite{fei2024videoccam}& 9B & \underline{58.5} & 3.98 & \textbf{42.9}&3.57   &50.3 &52.4 &35.6&\underline{69.0} &43.1 & - &\textbf{64.6} \\
Long-LLaVA~\cite{wang2024longllava}& 13B & - & - &- &- &  51.9 &-  & \underline{52.1}  &- &- &- &- \\
\midrule
\rowcolor{ModelGreen}\textbf{Video-XL} & 7B & \textbf{64.9} & \textbf{4.50} & \textbf{45.5} &\textbf{4.21}    &\textbf{55.5} &\textbf{61.0}  &\textbf{61.6}    & \textbf{70.6}  &\textbf{50.7}  &   \underline{3.17} &55.3 \\ 
\bottomrule
\end{tabular}}
\vspace{-2mm}
\caption{Experimental results on mainstream video benchmarks. ``LongVideo." and ``VideoChat." refer to LongVideoBench and VideoChatGPT Bench, respectively. $\dag$ indicates that the results on VNBench and LongVideoBench were reproduced using their official weights.} 
\vspace{-2mm}
\label{tab:mlvu} 
\end{table*}

\textbf{Composite Data curation}. Long-video instruction tuning data is very scarce in reality, which hinders the effective training of Video-XL. To mitigate this problem, we propose the composite curation of training data, where extra data resources are introduced to enhance the training effect. 

First, considering that understanding visual information in an image is foundational to video comprehension, we employ image captioning and QA data for augmentation. To facilitate knowledge transfer, we define a unified pipeline to transform all data into a uniform format. Specifically, we regard an arbitrary input data instance, whether it's a single-image, a multi-image, or a video, as a super image. We then divide the super image into multiple patches, each one in a resolution of $336 \times 336$. For each patch, we make use of CLIP to encode it as $M$ visual tokens ($M = 144$ in our implementation). The image captioning and QA data is relatively abundant in reality. In our work, the following datasets are collected: Bunny~\cite{he2024bunny}, Sharegpt-4o~\cite{sharegpt4o} (57k), and MMDU~\cite{liu2024mmdu} (20k). These datasets are combined with our video data, which contains NExT-QA~\cite{xiao2021next} (32k), Sharegpt-4o~\cite{sharegpt4o} (2K), CinePile~\cite{rawal2024cinepile} (10k), VCG~\cite{Maaz2024VideoGPT+} (25k) and in-house video captions with GPT-4V (11k). 

% Additionally, we curate 20k video samples from our proposed VICO, which will be discussed in the following part.   

% To mitigate the scarcity of high-quality long video data, we first explore the knowledge transfer phenomenon in long video understanding by establishing a unified multimodal data encoding protocol. Specifically, given a single image $SI$ with height $h$ and width $w$, we divide it into multiple patches, each with a resolution of $336 \times 336$ which will be encoded by CLIP separately. In contrast, a multi-image $MI$ or video $V$ containing $N$ frames will be treated as an extended image where each frame is considered a patch. In other words, the input visual signal $Z$ will be encoded by CLIP using $g_\phi(\cdot)$ parameterized by $\phi$:
% \begin{equation*}
% \varphi(Z) = 
% \begin{cases}
% (h // 336 \times w // 336) \times M, & \text{if } Z \in SI \\
% N \times M, & \text{if } Z \in \{MI,V\}
% \end{cases}
% \end{equation*}

% Specifically, for image data, we use the Bunny dataset~\cite{he2024bunny}, 57k images from Sharegpt-4o~\cite{sharegpt4o}, and utilize 20k data points extracted from MMDU~\cite{liu2024mmdu}. For video data, we collect NExT-QA  32k~\cite{xiao2021next}, Sharegpt-4o 2K~\cite{sharegpt4o}, CinePile 10K~\cite{rawal2024cinepile}, VCG 25k~\cite{Maaz2024VideoGPT+}  and 11k in-house samples with GPT-4V annotations for video captions. Additionally, we curate 20k video samples from our proposed VICO, which will be discussed in the following part.   

Second, understanding long videos also relies on precise and comprehensive utilization of proper information from the input. Thus, we additionally curate another synthetic dataset, called Visual Clue Order (VICO), to strengthen this fundamental capability. VICO contains 20k QA pairs, each one is associated with a video of 3 minutes on average. The videos are sourced from CinePile~\cite{rawal2024cinepile}, which covers diverse genres, like movies, documentaries, games, sports, etc. As shown in Figure~\ref{fig:vico_pipe}, each long video is segmented into 14-second clips. For each clip, we use the VILA-1.5~\cite{lin2024vila} to generate detailed descriptions. Based on these captions, we leverage GPT-4 to extract the key events and arrange them in a temporal order. VICO requires models to identify and reason about key information in a long video, thereby enhancing their long video comprehension capabilities.

\section{Experiment}

\subsection{Implementation} 
% We implement Video-XL by using Qwen-2-7B~\cite{yang2024qwen2} as our backbone LLM. 

Video-XL is trained on Qwen-2-7B~\cite{yang2024qwen2}. We employ a two-stage strategy to train Video-XL. During pre-training, we use the Laion-2M dataset~\cite{he2024bunny} to optimize the projector, where visual embeddings from a CLIP-ViT-L \cite{radford2021clip} based vision encoder are aligned with the text embeddings of LLM. During fine-tuning, we apply visual instruction tuning to optimize the parameters of vision encoder, projector and LLM. The batch sizes for pre-training and finetuning are 8 and 1, while the learning rate is 5e-5 for pre-training
and 1e-5 for fine-tuning, with linear decay and no warmup.  All experiments are conducted on one 8$\times$A800-80GB machine. 

\begin{figure*}[t]
    \centering
    \vspace{-0.3cm}
    \includegraphics[width=0.9\textwidth]{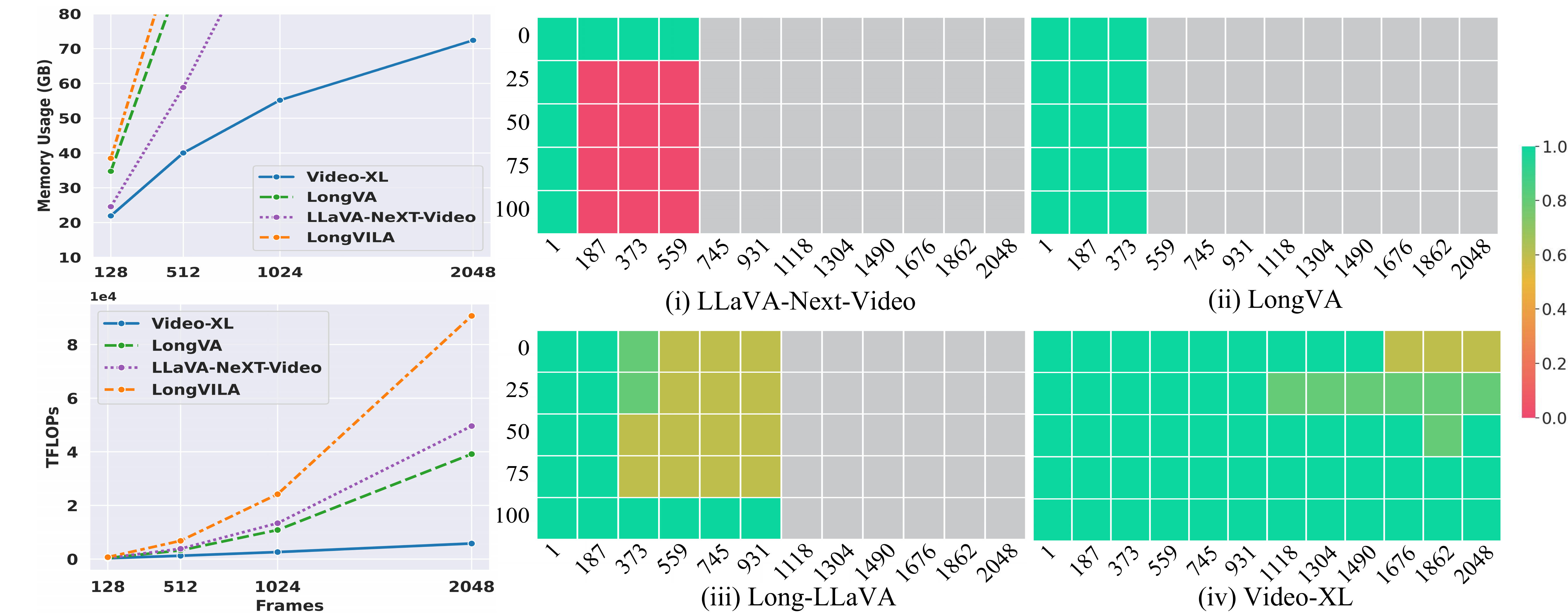}
    % \vspace{-0.5cm}
    \caption{(Left) Comparison of the memory usage and the forward FLOPs of different models. (Right) Results on the Needle-in-a-haystack evaluation within a single A100 80GB GPU. The x-axis represents the total number of frames in the video haystack. The y-axis shows the position where the needle image is located. Gray grids mean ``OOM'.  }
    \label{fig:needle}
    \vspace{-0.3cm}
\end{figure*}

\subsection{Benchmarks}
We empirically evaluate the effectiveness of Video-XL based on several popular long video understanding benchmarks. 1. MLVU~\cite{zhou2024mlvu}, a comprehensive benchmark which is made up of both multiple choice and generation tasks.
2. Video-MME~\cite{videomme}, another extensive benchmark covering videos of diverse genres and lengths (short, medium, and long).
3. VNBench~\cite{vnbench}, a synthetic benchmark focused on assessing models’ ability to handle long-video tasks, such as retrieval, ordering, and counting.
4. LongVideoBench~\cite{wu2024longvideobench}, a benchmark designed for tasks that require precise retrieval and reasoning over detailed multi-modal information within extended inputs.
5. VideoVista~\cite{li2024videovista}, which aims to evaluate a model's long-context reasoning ability over videos of varying durations.
In addition to the above long video evaluation, we also make extension for two short video question answering benchmarks: the VideoChatGPT Benchmark\cite{maaz2023videochatgpt} and MVBench\cite{li2024mvbench}.

\subsection{Main Results}
We present the performance of Video-XL on popular long-video benchmarks in Table~\ref{tab:mlvu}. 
Our results show that Video-XL consistently achieves strong performances across these experiments. 
Notably, it outperforms the existing methods on both Dev and Test tasks of MLVU. It even surpasses GPT-4o on the Dev tasks despite having only 7B parameters. 
For Video-MME, Video-XL achieves accuracies of 55.5\% and 61.0\% for the `without' and `with-subtitle' settings, respectively, which yields competitive results compared to the state-of-the-art models on this benchmark. 
For VNBench, Video-XL sets the top performance among the open-source models, leading the previous best model by nearly 10\% in accuracy. Once again, it surpasses GPT-4V and achieves a comparable performance to GPT-4o in this evaluation. 
While for VideoVista, Video-XL ranks the first place among all open-source MLLMs, trailing only behind GPT-4o and Gemini-1.5~\cite{reid2024gemini}. 
Video-XL also brings forth the highest performance among all open-source models with no more than the 7B parameters on the Dev task of LongVideoBench. 
Last but not least, although designed primarily for long video understanding tasks, Video-XL excels in short video tasks as well, yielding competitive results on both VideoChatGPT and MVBench benchmarks.

\begin{figure}[t]
    \centering
    \begin{minipage}[t]{0.5\textwidth}
        \centering
        \vspace{-0.1in}
        \resizebox{0.9\linewidth}{!}{
        \renewcommand{\arraystretch}{1.15}
        \begin{tabular}{>{\kern-0.5\tabcolsep}l|cccc<{\kern-0.5\tabcolsep}}
            \toprule
            \textbf{Model} & \textbf{MLVU} & \textbf{VideoMME} & \textbf{MME} & \textbf{MMB} \\
            \midrule
            Pooling  & 33.7 & 41.0 & 1405.5 & 62.3 \\
            Q-Former  & 35.1 & 42.1 & 1410.2 &61.9 \\
            LLaMA-VID  & 35.5 & 45.7 &1421.2 &64.3 \\
            LLaMA-Adapter  & 35.3 &42.2 & 1418.3& 65.5 \\
            C-Abstractor  & 37.1 & 46.3 & 1440.2 & 65.1 \\
            \midrule
            \rowcolor{ModelGreen}Video-XL  & 41.4 & 52.0 & 1510.2  & 70.9 \\
            \midrule
            Upper-bound & 41.8 & 52.6 & 1533.7  &71.6 \\
            \bottomrule
        \end{tabular}
        }
        \vspace{-0.08in}
        \captionof{table}{Comparison of compression techniques. All methods are implemented in the same setting and conducted with 16$\times$ compression.} 
        \label{tab:compression}
    \end{minipage}%
    \hfill
    \begin{minipage}[t]{0.45\textwidth}
        \centering
        \includegraphics[width=0.9\textwidth]{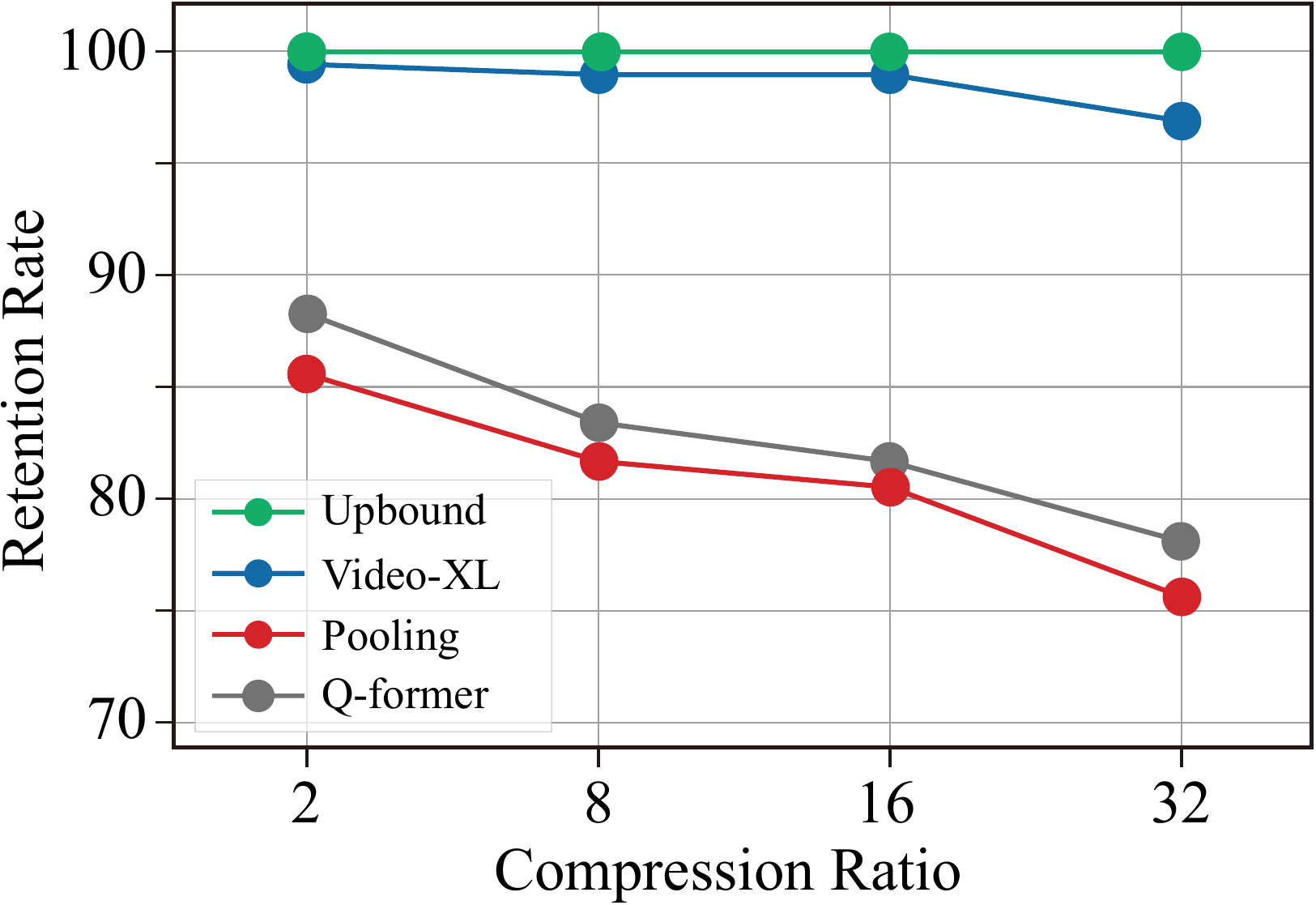}
        \vspace{-5pt}
        \caption{MLVU performance with variant compression ratios. The retention rate is calculated as the ratio to the upper-bound.}
        \vspace{-10pt}
        \label{fig:compression_ratio}
    \end{minipage}
\end{figure}

% \begin{table}[t]
%     \centering
  
%     \vspace{-0.1in}
%     \resizebox{0.9\linewidth}{!}{
%     \renewcommand{\arraystretch}{1.15}
%     \begin{tabular}{>{\kern-0.5\tabcolsep}l|cccc<{\kern-0.5\tabcolsep}}
%         \toprule
%         \textbf{Model} & \textbf{MLVU} & \textbf{VideoMME} & \textbf{MME} & \textbf{MMB} \\
%         \midrule
%         Pooling  & 33.7 & 41.0 & 1405.5 & 62.3 \\
%         Q-Former  & 35.1 & 42.1 & 1410.2 &61.9 \\
%         LLaMA-VID  & 35.5 & 45.7 &1421.2 &64.3 \\
%         LLaMA-Adapter  & 35.3 &42.2 & 1418.3& 65.5 \\
%         C-Abstractor  & 37.1 & 46.3 & 1440.2 & 65.1 \\
%         \midrule
%         \rowcolor{ModelGreen}Video-XL  & 41.4 & 52.0 & 1510.2  & 70.9 \\
%         \midrule
%         Upbound & 41.8 & 52.6 & 1533.7  &71.6 \\
%         \bottomrule
%     \end{tabular}
%     }
%     \vspace{-0.12in}
%       \caption{Comparison with previous visual compression techniques. All methods are retrained in a 16x compression.}
%     \label{tab:compression}
% \end{table}

% \begin{figure}[t]
% \centering
% \includegraphics[width=0.4\textwidth]{figs/compression.pdf}
%     \caption{Evaluation on MLVU with various compression ratios. The compression retention rate is calculated as the ratio of compressed results to the upper bound.}
%     \vspace{-5pt}
%     \label{fig:compression}
% \end{figure}

\subsection{Extra-Long Evaluation}
To explore Video-XL's ability to process extra-long video inputs, we further conduct the Needle-In-The-Haystack evaluation~\cite{longva} based on an A100-80GB GPU. We consider two types of baselines in our evaluation: 1) LLaVA-NexT-Video and LongLLaVA, which rely on position extrapolation methods to make extension for longer inputs, and 2) LongVA, which fine-tunes the MLLM to handle longer inputs. As shown in Figure~\ref{fig:needle}, Video-XL exhibits notable advantages over the baselines. First, Video-XL is able to cover much longer video inputs. However, neither LLaVA-NexT-Video nor LongLLaVA can support more than 1000 frames due to the constraint of computation cost, while LongVA is only fine-tuned to support less than 400 frames. Second, Video-XL well maintains a superior performance. It preserves 100\% accuracy within 128 frames, which is the maximum length of its fine-tuning data; meanwhile, it achieves nearly 95\% accuracy when dealing with longer inputs. In contrast, LLaVA-NexT-Video and LongLLaVA suffer from inferior retrieval performance, while LongVA can only handle inputs within its fine-tuned length.

% To demonstrate Video-XL's ability to process extremely long contexts, we conduct a needle-in-the-haystack evaluation~\cite{longva} on an A100 80GB GPU, as shown in Figure~\ref{fig:needle}. It is evident that both LLaVA-NexT-Video and LongLLaVA use a naive position encoding extrapolation algorithm but struggle to capture fine-grained information when provided with more input context. LongVA fine-tunes the LLM to handle extended input lengths but still incurs high computational costs, limiting it to processing approximately 400 frames under constrained computational resources. In contrast, benefiting from a high-fidelity 16$\times$ compression ratio, Video-XL can process 2,048 frames and achieves nearly 95\% accuracy. This suggests that our model strikes an optimal balance between accuracy and computational efficiency.

\subsection{Inference Efficiency}
We further evaluate the inference efficiency of Video-XL in comparison to three baselines: LongVA, LLaVA-NeXT-Video, and LongVILA. As shown in Figure~\ref{fig:needle} (left), Video-XL significantly reduces memory usage thanks to its compression of visual information. The substantial reduction in memory consumption allows it to process over 2048 frames with a single A100-80GB GPU. In addition, Video-XL also results in much smaller TFLOPs than the baseline methods, as it eliminates the need for direct self-attention over long input sequences.

% \begin{figure*}[t]
%     \centering
%     \vspace{-0.3cm}
%     \includegraphics[width=0.9\textwidth]{figs/data.pdf}
%     % \vspace{-0.5cm}
%     \caption{(Left) Performance of Video-XL trained on different data. (Right) The effectiveness of VICO data.}
%     \label{fig:data_mix}
%     \vspace{-0.3cm}
% \end{figure*}

\subsection{Ablation Studies}
\label{sec:ablation}
We conducted extensive ablation studies to explore Video-XL's effectiveness regarding its compression mechanism, training method, and data curation. 

% We conducted extensive ablation studies on the proposed Video-XL, encompassing the design of the compression mechanism and the exploration of knowledge transfer.

\textbf{Compression mechanism}. First, we compare Video-XL with previous common pre-compression methods, including average pooling, Q-Former~\cite{li2023blip2}, LLaMA-VID~\cite{llamavid}, LLaMA-Adapter~\cite{gao2023llamaadapterv2}, and C-Abstractor~\cite{cha2024honeybee}. For a fair comparison, these methods are implemented based on their official codes, but switched to the same architecture and training data as our method. With a uniform compression ratio of 16$\times$, we report the results on two long video benchmarks, MLVU-test and VideoMME, as well as two popular VQA benchmarks, MME~\cite{fu2023mme} and MMB~\cite{liu2023mmbench}. As shown in Table~\ref{tab:compression}, Video-XL significantly outperforms previous methods across all benchmarks, particularly on long video benchmarks, which require fine-grained detail understanding and long-term relational reasoning. Additionally, Video-XL achieves high-fidelity compression with minimal performance loss, even at compression ratios as high as 16$\times$. Moreover, we further explore the performance under various compression ratios (2$\times$, 8$\times$,16$\times$, 32$\times$) in Figure~\ref{fig:compression_ratio}. Note that 32$\times$ is directly tested without fine-tuning. In these experiments, Video-XL maintains a close performance as the upper-bound, outperforming the baselines by a large margin. Meanwhile, it also effectively preserves its performance for the unseen compression ratio (32$\times$), suggesting the generality of the proposed method.

% since Video-XL is trained to support various compression ratios, we expand the compression ratio to 32× during training and evaluate its compression quality under different ratios in Figure~\ref{fig:compression_ratio}. Video-XL maintains top accuracy across all compression ratios, outperforming most compression baselines by a large margin. Overall, we recommend a 16× compression ratio, as it preserves most information with high efficiency.

\begin{table}[t]
 \centering
    \vspace{-0.1in}
    \resizebox{0.9\linewidth}{!}{
    \renewcommand{\arraystretch}{1.15}
    \begin{tabular}{>{\kern-0.5\tabcolsep}cc|cccc<{\kern-0.5\tabcolsep}}
        \toprule
        \textbf{Train} & \textbf{Test} &\textbf{MLVU} & \textbf{VideoMME} & \textbf{MME} & \textbf{MMB} \\
        \midrule
        \crossmark & \crossmark & 39.8 & 50.9 &1460.6 &70.9 \\
        \crossmark & \checkmark & 39.6 & 50.8 & 1455.0 &70.8 \\
        \checkmark & \crossmark & 41.5 & 52.0 & 1515.5 &71.2  \\
        \rowcolor{ModelGreen}\checkmark & \checkmark & \textbf{41.6} & \textbf{52.3} &\textbf{1520.0} &\textbf{71.3} \\
        \bottomrule
    \end{tabular}
    }
    \vspace{-0.1in}
    \caption{Evaluation of dynamic compression strategy. }\label{tab:dynamic_compression}
\end{table}

\begin{table}[t]

    \vspace{-0.1in}
    \resizebox{0.98\linewidth}{!}{
    \renewcommand{\arraystretch}{1.15}
    \begin{tabular}{>{\kern-0.5\tabcolsep}c|cccc<{\kern-0.5\tabcolsep}}
        \toprule
        \textbf{Settings}  &\textbf{MLVU} & \textbf{VideoMME} & \textbf{MME} & \textbf{MMB} \\
        \midrule      
        w/o random compre. & 40.5 & 51.0 &1500.4 &70.3 \\
        w/o curriculum learn. & 41.1 & 51.6 &1512.4 &71.0 \\
        \rowcolor{ModelGreen}Ours & \textbf{41.6} & \textbf{52.3} &\textbf{1520.0} &\textbf{71.3} \\
        \bottomrule
    \end{tabular}
    }
    \vspace{-0.1in}
     \caption{Evaluation of curriculum learning.}
     \label{tab:random_compression}
\end{table}

\begin{table}[t]
    \centering
    \begin{minipage}[t]{0.48\textwidth} % 设置表格的宽度为总宽度的48%
        \vspace{-0.1in}
        \resizebox{\linewidth}{!}{
            \renewcommand{\arraystretch}{1.15}
            \begin{tabular}{>{\kern-0.5\tabcolsep}c|c|c|cccc<{\kern-0.5\tabcolsep}}
                \toprule
                \textbf{Video}  &\textbf{Single Image} & \textbf{Multi Image} & \textbf{TR} & \textbf{NQA} & \textbf{AO} & \textbf{Avg} \\
                \midrule      
                100k & - & - &73.4 &64.5 & 53.6 &63.8 \\
                100k & 350k  & - &77.5 &66.9 &54.0 &66.1 \\
                100k & 700k  & - &80.6 &70.0 &54.1 &68.2 \\
                100k & 1M  & - &81.3 &69.8 &53.8 &68.3 \\
                 \rowcolor{ModelGreen}100k & 700k  & 20k &\textbf{82.0} &\textbf{70.3} &55.3&\textbf{69.5} \\
                100k & 700k  & 40k &82.1 &70.1 &\textbf{55.4} &69.2 \\
                \bottomrule
            \end{tabular}
        }
        \vspace{-0.1in}
        \caption{Analysis of training effect from different data.}
        \label{tab:data_mix}
    \end{minipage}
\end{table}
\begin{figure}[t]
    \centering
    \begin{minipage}[t]{0.48\textwidth} % 设置图像的宽度为总宽度的48%
        \centering
        \includegraphics[width=0.7\linewidth]{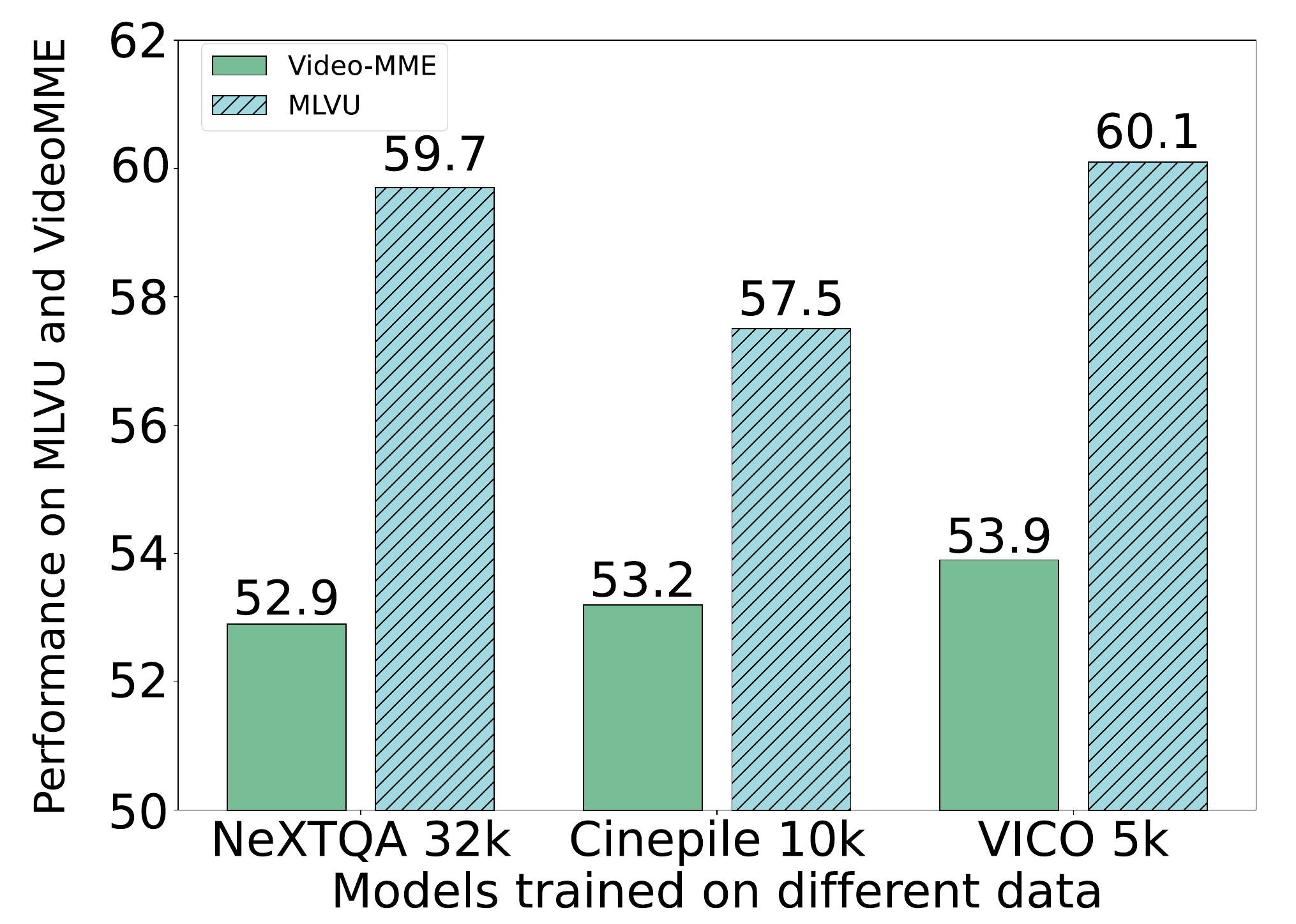} % 在此处替换为图像文件路径
        \caption{Analysis of training effect from VICO.}
        \vspace{-10pt}
        \label{fig:vico}
    \end{minipage}
\end{figure}

\textbf{Dynamic compression strategy}. Second, we analyze the effect of dynamic compression strategy. In this experiment, we compare the settings where dynamic compression is disabled, or individually enabled for training and testing. If dynamic compression is disabled, we perform fixed compression based on an interval of 1440 tokens. From the experiment results in Table~\ref{tab:dynamic_compression}, we can validate the effectiveness of our method, as it leads to a substantial improvement over the dynamic-compression disabled baselines. Meanwhile, we can also observe that it's necessary to conduct dynamic compression during training, as no improvements are obtained if it's only enabled for testing stage. More discussions on this issue are provided in the supplementary.

% We validate our proposed dynamic compression strategy by applying it individually during training and testing, as well as in both stages. The results are shown in Table~\ref{tab:dynamic_compression}. First, we perform fixed compression by setting the interval size to 1440 and testing with both fixed and dynamic settings. However, the model with dynamic compression achieves inferior results, likely due to the loss of information from the massive video QA data during compression. To address this, we retrained models using dynamic compression, which yielded promising results on both video and image benchmarks, suggesting that dynamic compression can preserve more fine-grained information. Further discussion on this topic is provided in the supplementary.

\begin{figure*}[t]
    \centering
\includegraphics[width=0.8\textwidth]{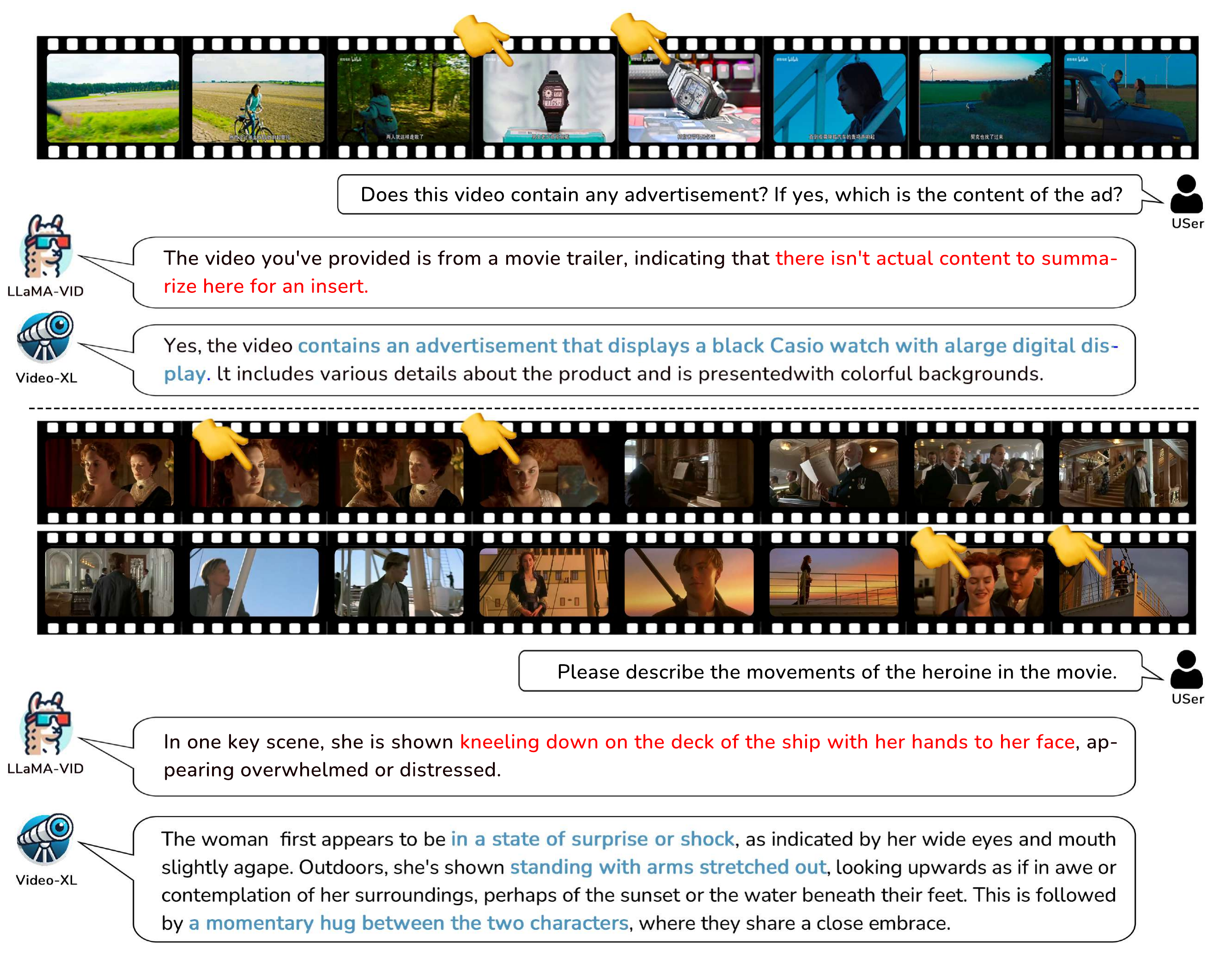}
    % \vspace{-2pt}
    \caption{Qualitative evaluation of Video-XL in two tasks.}
    % \vspace{-10pt}
    \label{fig:vis}
\end{figure*}

% 每个chunk包含的信息密度是不同的，因此需要不同的压缩比。同时大的压缩比对信息损害大，导致训练难度增加；而小压缩比的训练较为容易。
\textbf{Curriculum learning}. To assess the effectiveness of curriculum learning, we re-train the model with two settings: 1. using randomized compression ratios within 16$\times$, i.e., w/o curriculum, 2. using a fixed compression ratio 16$\times$, i.e. w/o random (16$\times$ is the compression ratio used for testing). As shown in Table~\ref{tab:random_compression}, the our methods substantially improves upon the two baselines, indicating the necessity to learn progressively from small compression ratios.

% the model is more capable of perceiving details in long contexts with the aid of curriculum learning. Additionally, we conduct an empirical study with random compression ratio settings, which also enhances the model's ability to learn compression functionality.

% The improvement of the training process for Video-XL is a significant technical factor, which includes the chunk-wise compression ratio control and curriculum learning-based training methods. The results are shown in Table~\ref{tab:random_compression}. Firstly, we replace the chunk-wise random compression ratio with a fixed and dedicated one, where each training instance is assigned a specific compression ratio equally. We observe that the chunk-wise setting facilitates better learning of the compression functionality. Secondly, based on the chunk-wise compression ratio control, we retrain the model without utilizing the curriculum learning method. In this case, the compression ratio during training is independent of the training steps. However, this approach leads to significant visual information loss.

\textbf{Composite data curation}. To explore the effect from different data sources (video, single-image, multi-image), we make fine-grained analysis based on three types of tasks from MLVU: 1. Topic Reasoning (TR): for holistic understanding capability, 2. Needle QA (NQA): for single-detail understanding capability, 3. Action Order (AO): for multi-detail understanding capability. As shown in Table~\ref{tab:data_mix}, the increasing of image data effectively enhances the model's holistic (TR) and single-detail (NQA) capability, however, it contributes little to multi-detail (AO) capability (from 1st row to 3rd row). Meanwhile, once sufficient image data is presented, the additional benefit becomes marginal (as reflected from the 3rd row to the 4th row). Finally, the introduction of multi-image data significantly improves the model’s multi-detail capability, as it enables the model to learn fine-grained relationships within long inputs. The above observations indicate that different data sources are complementary to each other, which jointly contribute to the superior performance of Video-XL.

% First, we standardize single and multi-images to align with the structure of video frames, facilitating knowledge transfer for long video understanding. To further investigate this, we use data of varying scales and proportions to train models, and report results on specific MLVU subtasks: Topic Reasoning (TR), Needle QA (NQA), and Action Order (AO). As shown in Table~\ref{tab:data_mix}, increasing the amount of image data effectively enhances the model's holistic (TR) and single-detail (NQA) capabilities, with minor improvements in multi-detail capabilities (from line 1 to line 3. However, once the image data reaches a certain threshold, the benefits become minimal (as seen from line 3 to line 4). The introduction of multi-image data significantly improves the model’s multi-detail capabilities by enabling fine-grained comparisons within longer sequences. Consequently, we use a training dataset that combines single images, multi-images, and videos, yielding strong long video understanding capabilities and efficient training. 

To further analyze the effect of VICO dataset, we re-train the model using three video instruction-tuning datasets: (a) NeXTQA 32k, (b) CinePile 10k, and (c) VICO 5k. The corresponding results on Video-MME and MLVU are shown in Figure~\ref{fig:vico}. Although VICO is the smallest of all datasets, it substantially outperforms the other two datasets which contain more training samples (5k vs. 32k and 10k), demonstrating its value to establish the long-video understanding capability for MLLMs. We also discuss the effect from scaling up VICO in our supplementary material.

\subsection{Qualitative Evaluation} % 用 subsection ，写长点，12行以上，写具体点，对比其他方法
We leverage qualitative evaluation for an intuitive analysis of Video-XL. In this experiment, we make comparison with LLaMA-VID \cite{llamavid} based on extra-long videos (over 30 minutes). As shown in Figure~\ref{fig:vis}, Video-XL accurately locates the inserted advertisement and presents its details; in contrast, LLaMA-VID struggles to comprehend the video and make judgment on whether an advertisement is inserted. Additionally, Video-XL effectively summarizes the plots about the heroine in the long video, whereas LLaMA-VID only returns a short and inaccurate description. We include more qualitative analysis in our supplementary material.

% As shown in Figure~\ref{fig:vis}, we present some qualitative results of Video-XL on long videos (over 30 minutes) and compare to the popular model LLaMA-VID. As shown in the figure, LLaMA-VID has difficulty in comprehending the long videos and identifying the clues whether an advertisement is inserted. In contrast, our Video-XL accurately locates the inserted advertisement and presents the details. Moreover, Video-XL well summarizes the plots about the heroine in the long video, which well follows user's instruction, while LLaMA-VID gives short and inaccurate description. Additional samples are provided in the Supplementary.

% , which compresses token counts to enable long video understanding. 
% For instance, it struggles to comprehend detailed clues in a long video and reason over extended contexts. In contrast, Video-XL better preserves fine-grained information. It not only identifies the inserted ad in the video but also accurately describes its color and on-screen text. For multi-detail questions, it establishes long-range reasoning relationships involving the targeted character. Additional samples are provided in the Appendix.

\section{Conclusion}
In this paper, we introduce Video-XL, which enables the processing of long videos on top of the compressed representations generated by our visual summarization token (VST). To better retain the visual information, we conduct dynamic compression based on the information density of the video. Additionally, to optimize the training effect, we design a curriculum learning method, allowing for progressive learning of different compression ratios. We also propose composite data curation, which jointly utilizes multiple data sources to improve the model's performance. The effectiveness of Video-XL is empirically verified, as it achieves superior performance across popular long-video benchmarks and delivers competitive compression quality and cost-effectiveness in our experiments.

% In this paper, we introduced Video-XL, designed for extremely long video understanding. Video-XL fully excavates the potential of LLM to compress massive visual tokens. To better retain fine-grained information, we propose a dynamic compression strategy, which customizes compression granularity based on the information density of different video intervals. To facilitate the training process, we utilize curriculum learning and employ a composite data curation to mitigate the scarcity of long video data. The proposed Video-XL achieves leading performance in several popular video benchmarks, and demonstrates a promising balance between effectiveness and efficiency. 

{
    \small
    \bibliographystyle{ieeenat_fullname}
    \bibliography{main}
}


\clearpage

\setcounter{page}{1}
\maketitlesupplementary

\appendix % 从这里开始编号重新计数
\setcounter{figure}{0} 
\setcounter{table}{0}

\section*{Overview of Supplementary Material}

\begin{itemize}
  \item  \ref{appendix:limitation}: \textbf{Limitations and Future Works}
   \item  \ref{appendix:relation}: \textbf{Relations to Concurrent Works}
    \item  \ref{appendix:llm_compressor}: \textbf{Details in LLM Compressor}
    \item \ref{appendix:implement}:
\textbf{Further Discussion of Video-XL }
    \item \ref{appendix:vico}: \textbf{Analysis of VICO Dataset}
    \item \ref{appendix:experiment}: \textbf{Experimental Settings and Additional Results}
    \item \ref{appendix:qualitative}: \textbf{Qualitative Results}

\end{itemize}

\section{Limitations and Future Works}
 \label{appendix:limitation} 

Although Video-XL has a strong capacity to handle extremely long videos, it also has several limitations: (i) Large training memory cost: During training, we unfreeze all parameters of CLIP, the vision-language projector, and the LLM, requiring substantial GPU memory. Additionally, processing video frame extraction and handling a large number of visual tokens demand extra computational resources. We will continue to optimize the training process to improve efficiency or integrate smaller-scale LLMs. (ii) Performance decline with increasing visual tokens: As shown in the Needle-in-the-haystack evaluation, Video-XL occasionally makes errors when the context exceeds 1,000 frames. 
% We argue that the model still suffers from information decay due to token aggregation. 
In the future, we will continue to improve the visual compression module and reduce the information decay for longer video understanding.

\section{Relation to Concurrent Works}
 \label{appendix:relation} 
In this section, we compare and discuss the relation between our Video-XL with the concurrent works including LongVA~\cite{longva} and VoCo-LLaMA~\cite{ye2024voco}.

\textbf{Comparison to LongVA.} 
Though Video-XL shares a similar model architecture with LongVA, they have several important distinctions. First, although the unified visual encoding mechanism resembles that of LongVA, Video-XL simultaneously encodes single images, multiple images, and videos during training, whereas LongVA only uses image data. Furthermore, we explore the knowledge transfer extensively, conducting experiments to demonstrate that single images and multiple images contribute to long video understanding from different perspectives. Regarding technical contributions to improving long-context processing capability, LongVA fine-tunes the LLM to extend its context length, while Video-XL designs Visual Summarization Token and learns to compress visual tokens. Consequently, Video-XL can handle more visual tokens than LongVA using the same computing device.

\textbf{Comparison to VoCo-LLaMA.} Like VoCo-LLaMA, Video-XL also leverages the inherent capability of LLMs to compress visual tokens. Beyond VoCo-LLaMA for image token compression, Video-XL introduces several significant technical contributions that distinguish it from that one. 
Firstly, VoCo-LLaMA appends all special tokens at the end of a chunk, while our Video-XL splits long visual token sequences into fine-grained intervals and interleaved special tokens. Thus, VoCo-LLaMA has difficulty in handling long videos.
Secondly, the dynamic compression is utilized in Video-XL to ensure the control of compression granularity within a video.
Thirdly, we optimize the training process by utilize composite data curation and curriculum learning techniques.
% Firstly, the LLM compression mechanism in VoCo-LLaMA is designed for images rather than that designed for videos in Video-XL. Specifically, we utilize the dynamic chunk adjustor to ensure the consistency of visual content within a video chunk and propose a progressive chunk-wise compression strategy during the training process. . 
% Fourth, to manage the imbalance of visual redundancy, we  
% Secondly, VoCo-LLaMA appends all special tokens at the end of a chunk, while our Video-XL splits long visual token sequences into fine-grained clips and interleaved special tokens. Thus, VoCo-LLaMA has difficulty in handling long videos.
% , as the generated tokens may exceed the context length of the LLM. 
% In contrast, Video-XL introduces several technical designs for realizing long video understanding. Specifically, to address the attention bias caused by incoherent visual content in a chunk, 
% Lastly, s
Since the official weights of VoCo-LLaMA have not been released, we cannot comprehensively compare the models on long video benchmarks. Therefore, we report our performance (in Table~\ref{voco}) on image understanding benchmarks for reference, though our model mainly designed is for video understanding. It shows that Video-XL achieves significantly better results than VoCo-LLaMa even though on image benchmarks.
 % and retains better information after compression

\begin{table}[h]
\centering
    \centering
    \vspace{-0.1in}
    \renewcommand{\arraystretch}{1.15}
    \resizebox{0.98\linewidth}{!}{
    \begin{tabular}{>{\kern-0.5\tabcolsep}lc|ccc<{\kern-0.5\tabcolsep}}
        \toprule
        \textbf{Model} & \textbf{Compression Ratio} & \textbf{MMB} & \textbf{GQA} & \textbf{SEED} \\
        \midrule
        VoCo-LLaMA & - & 64.0 & 61.1  &57.9 \\
        VoCo-LLaMA & 8 & 60.5 & 60.4 & 56.3 \\
        VoCo-LLaMA &16 & 59.4 & 60.2 & 56.2 \\
        \midrule
        \rowcolor{ModelGreen}Video-XL  & - & 71.6 & 60.0  &61.6 \\
        \rowcolor{ModelGreen}Video-XL  & 8 & 71.4 & 59.3  & 61.2 \\
        \rowcolor{ModelGreen}Video-XL  & 16 & 70.9 & 59.1  & 61.0 \\
        % \rowcolor{cyan!15}Video-XL mix & 41.6 & 52.3 & 1520.0  & 71.3 \\
        \bottomrule
    \end{tabular}}
     \caption{Comparison with VoCo-LLaMA under different compression ratios on image understanding benchmarks.}
    \label{voco}
\end{table}

\section{Details in LLM Compressor}
 \label{appendix:llm_compressor} 
In our work, we split long visual sequences into shorter intervals and introduce the special tokens, namely the VSTs, which condense
LLM’s raw activations into more compact ones. Consequently, the same context window can intake more information from the previous context, which will benefit the prediction of new tokens. 
In each decoding layer of the LLM, let $D$ denote the LLM’s hidden size and $L$ is the size of the chunk, the input hidden states of VSTs ($H_{vst}\in\mathbb{R}^{k\times D}$) are transformed to query the raw KV activations within the chunk: $\{K,V\mid K\in \mathbb{R}^{L\times D}, V\in\mathbb{R}^{L\times D}\}$, where the condensed activations can be produced. Formally, 
\begin{gather}
    Q_{vst}\leftarrow H_{vst}W'_Q,\quad K_{vst}\leftarrow H_{vst}W'_K,\quad V_{vst}\leftarrow H_{vst}W'_V \\
    A \leftarrow \mathrm{softmax}\big( \mathrm{mask}(Q_{vst}\{K\oplus K_{vst}\}^T) \big) \\
    \quad V_{vst}\leftarrow A\{V\oplus V_{vst}\}^T,\quad O_{vst}\leftarrow V_{vst}{W'_O}^T.
\end{gather}
The newly generated KV activations for the VSTs, i.e., $K_{vst}, V_{vst} \in \mathbb{R}^{k\times D}$, which leads to a condensing ratio of $\alpha = L/k$ ($k \ll L$). Moreover, the VSTs are parameter-efficient because they primarily rely on the LLM's original parameters, introducing only a few additional projection matrices. For instance, they add no more than 1B parameters to the Qwen2 7B base model.

\section{Further Discussion of Video-XL}
 \label{appendix:implement} 

In this section, we discuss more details about Video-XL, including the training method, computational efficiency and generalization.

\textbf{Ablation studies on dynamic compression strategy.} Our dynamic compression strategy enables Video-XL to control the granularity of compression. We conduct empirical experiments to demonstrate the effectiveness of this strategy. Specifically, we compare dynamic compression with fixed methods that use different fixed interval sizes to encode video segments. As shown in Table~\ref{tab:appen_dynamic}, the performance of Video-XL is sensitive to the hyperparameter of fixed size $L$. Generally, larger $L$ results in poorer performance, as coarse-grained compression can damage detailed visual information. While smaller $L$ improves performance, it is highly benchmark-dependent and incurs longer training times. In contrast, dynamic compression allows Video-XL to achieve consistent performance across all benchmarks.

\begin{table}[h]
\centering
    \centering
    
    \vspace{-0.1in}
    \renewcommand{\arraystretch}{1.15}

    \begin{tabular}{>{\kern-0.5\tabcolsep}l|ccc<{\kern-0.5\tabcolsep}}
        \toprule
        \textbf{L (tokens)}  & \textbf{Video-MME} & \textbf{MLVU} & \textbf{MMB}  \\
        \midrule
         144 $\times$ 2  &41.3 & 52.3  &71.5 \\
         144 $\times$ 4  &40.7 & 52.1  &70.6 \\
        144 $\times$ 8  & 39.4 & 51.0  &69.6  \\
         144 $\times$ 16  & 38.9 & 50.5  &69.5  \\
          144 $\times$ 32  & 38.6 & 50.1  &69.2  \\
          \midrule
          Dynamic &41.6 &52.3 &71.3 \\
        \bottomrule
    \end{tabular}
    \caption{Ablations on dynamic compression strategy.}
    \label{tab:appen_dynamic}
\end{table}

\textbf{Ablation studies on training methods.}
%为了更有效地训练Video-XL, 我们研究了多种训练方式。他们都是基于visual insturction tunning的训练方式，区别主要在于fine-tuning阶段。第一种两阶段方法中，首先，我们尝试不设置压缩参数，训练1个epoch。随后，我们固定CLIP，projector以及之前训练好的LLM所有参数，只优化压缩中新引入的参数再用相同的数据训练1epoch。在第二种一阶段方法中，我们打开所有的参数用相同的训练数据训练2个epoch。我们在表中汇报了两种方法的结果，相比较于第二种方法，第一种方法拥有更优的训练效率，然而第二种方法的效果更加好。我们猜测这主要是由于projector提前对齐了CLIP和LLM，而无法适应压缩后的知识。
To train Video-XL more effectively, we explored various methods based on visual instruction tuning, primarily differing in the fine-tuning phase. In the first, two-stage method, we initially trained for one epoch without setting compression parameters. Next, we froze the parameters of CLIP, the projector, and all pre-trained LLM parameters, optimizing only the newly introduced parameters in the compression module for an additional epoch using the same data. In the second, single-stage method, we activated all parameters and trained for two epochs with the same training data. The results of both methods are reported in the table. Compared to the second method, the first method achieved better training efficiency, though the second method produced superior results. We speculate that this is mainly because the projector pre-aligns CLIP and the LLM, making it less adaptable to compressed knowledge.

\begin{table}[h]
\centering
    \centering
    
    \vspace{-0.1in}
    \renewcommand{\arraystretch}{1.15}
    \resizebox{0.98\linewidth}{!}{
    \begin{tabular}{>{\kern-0.5\tabcolsep}l|ccc|c<{\kern-0.5\tabcolsep}}
        \toprule
        \textbf{Model}  & \textbf{Video-MME} & \textbf{MLVU} & \textbf{MMB} & Training time \\
        \midrule
         Upper-bound  &41.8 & 52.6  &71.6 &- \\
        One-Stage  & 35.3 & 46.8  &66.7 & 1.5 days \\
        Two-stage  & 41.4 & 52.0 &70.9 &2 days \\
        \bottomrule
    \end{tabular}}
    \caption{Ablation studies on training methods of Video-XL.}
    \label{tab:training_method}
\end{table}

\textbf{Discussion on the computational efficiency.} Video-XL reduces the KV cache by $\alpha$ times where $\alpha$ is the \textit{average compression ratio} and hence the memory cost. This is because it only needs to store the compressed activations of the preceding chunks instead of the raw activations.
In terms of computation, the situation is a bit more complex. Specifically, Video-XL significantly reduces the computation in self-attention, because each token only needs to interact with local tokens within the chunk and preceding VSTs, which are approximately $\alpha$ times shorter than the raw context. However, it also triggers more computation to encode the inserted VSTs in other modules (e.g., MLP). Formally, given an LLM with a fixed number of layers, attention heads, and hidden size, let $s$ denote the input context length, $s^{pst}$ denote the cached context length, the forward FLOPs is:
\begin{equation}
    \mathrm{FLOPs} = F^{Att}(s, s^{pst}) + F^{Oth}(s),
\end{equation}
where $F^{Att}$ is the computation during self attention, and $F^{Oth}$ is the computation of other modules. For full-attention models, $s=n, s^{pst}=0$. For the LLM compressor in  Video-XL, the FLOPs is:
\begin{equation}
\resizebox{0.5\textwidth}{!}{$
    \mathrm{FLOPs}^{bcn}=\sum_{i=1}^{\lceil \frac{n}{w} \rceil}F^{Att}\left(\frac{(\alpha + 1)w}{\alpha}, \frac{(i-1)w}{\alpha}\right) + F^{Oth}(n+\lceil\frac{n}{\alpha}\rceil)
$}
\end{equation}
Specifically, denote the input sequence length as $s$, the cached sequence length as $s^{pst}$, query head number as $h^q$, key/value head number as $h^k$, the hidden size $D$, head dimension as $d$, intermediate size $I$, and vocabulary size $V$, FLOPS can be calculated as follows:

\begin{align}
    F^{Att} &= F^{qkv} + F^{qk} + F^{softmax} + F^{av} + F^{out} \nonumber\\
    F^{qkv} &= 2 \times s \times D \times d \times h^q + 2\times 2\times s \times D\times d\times h^k \nonumber\\
    F^{qk} &= 2 \times h^q \times s \times (s + s^{pst}) \times d \nonumber\\
    F^{soft} &= h^q \times (s + s^{pst}) \times (s + s^{pst}) \nonumber\\
    F^{av} &= 2 \times h^q \times s \times (s + s^{pst}) \times d \nonumber\\
    F^{out} &= 2 \times s \times d \times h^q \times D \\
    F^{Oth} &= F^{up} + F^{gate} + F^{down} + F^{lm} \nonumber\\
    F^{up} &= 2 \times s \times D \times 2 \times I \nonumber\\
    F^{gate} &= s \times \times I \nonumber\\
    F^{down} &= 2 \times s \times D \times I \nonumber\\
    F^{lm} &= 2 \times s \times D \times V
\end{align}

\begin{figure}[h]
    \centering
    \includegraphics[width=0.4\textwidth]{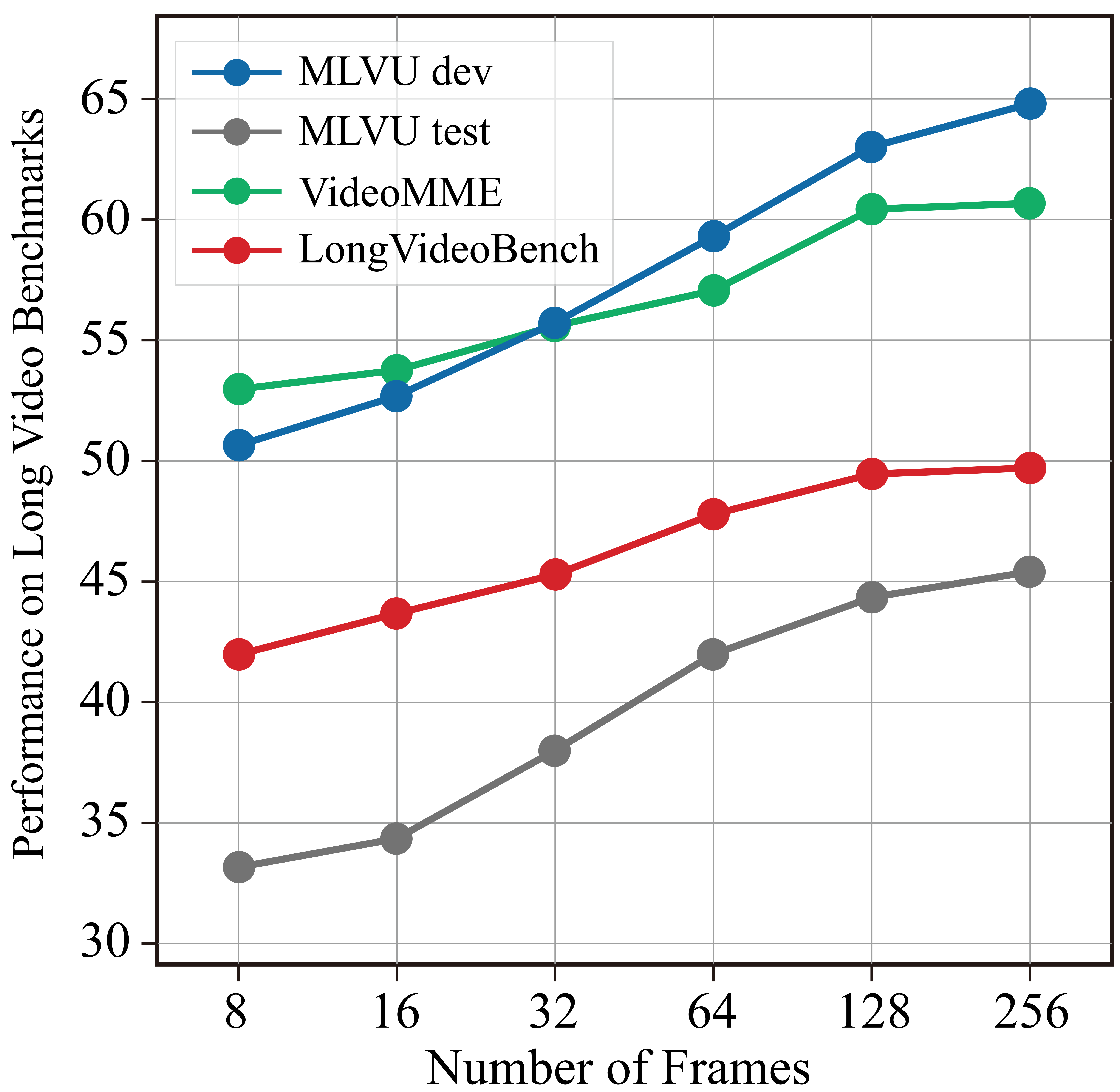}
    \caption{Video-XL can achieve better performance on long video benchmarks with increased context length.}
    \vspace{-5pt}
    \label{fig:length}
\end{figure}

\textbf{Discussion on the Generalization.}
% 语言模型；训练和推
%我们发现Video-XL具有很强的泛化性。一方面，语言压缩器可以灵活应用于各种语言模型上。除了Qwen2, 我们在Vicuna和LLaMA2上进行了实验，结果如下表所示。在未来，我们会引入参数更小的语言模型从而使得Video-XL可以在单张显卡上处理更长的视频。另一方面，Video-XL在训练时只需要用相对较短的视频（两分钟以下），但是可以在推理的时候处理接近一小时的视频。我们认为这主要归功于训练数据和模型的设计。在训练数据中，模型从图像，多图和短视频数据中学到了长视频理解相似的模式；而在模型的结构中，Video-XL对每个chunk中的visual tokens采用相对位置编码，因此在推理时候模型的理解能力可以泛化到无限长的视频。
We find that Video-XL demonstrates strong generalization capabilities. On one hand, the language compressor can be flexibly applied across various language models. In addition to Qwen2, we conducted experiments with Vicuna and LLaMA2, with results shown in the Table~\ref{tab:llm}. In the future, we plan to introduce smaller language models to enable Video-XL to process longer videos on a single GPU. On the other hand, Video-XL requires only relatively short videos (under two minutes) for training but can handle videos nearly an hour long during inference. We believe this is largely due to the design of both the training data and the model itself. In the training data, the model learns to understand long videos by leveraging images, multi-image sequences, and short video data. Furthermore, in the model’s architecture, Video-XL applies relative positional encoding to the visual tokens within each chunk, enabling it to generalize its understanding to infinitely long videos during inference. As shown in Figure~\ref{fig:length}, the increase of the context length can boost the performance of Video-XL on several long video benchmarks.

\begin{table}[h]
\centering
    \centering
   
    \vspace{-0.1in}
    \renewcommand{\arraystretch}{1.15}

    \begin{tabular}{>{\kern-0.5\tabcolsep}l|ccc<{\kern-0.5\tabcolsep}}
        \toprule
        \textbf{LLM}  & \textbf{Video-MME} & \textbf{MLVU} & \textbf{MMB}  \\
        \midrule
         Vicuna-7B  &36.5 & 48.1  &63.2 \\
        LLaMA2-7B  & 38.3 & 49.6  &66.8  \\
        Qwen2-7B  & 41.4 & 52.0 &70.9 \\
        \bottomrule
    \end{tabular}
     \caption{Video-XL has strong generalization to different LLMs.}
    \label{tab:llm}
\end{table}

\begin{figure}[h]
    \centering
    \includegraphics[width=0.45\textwidth]{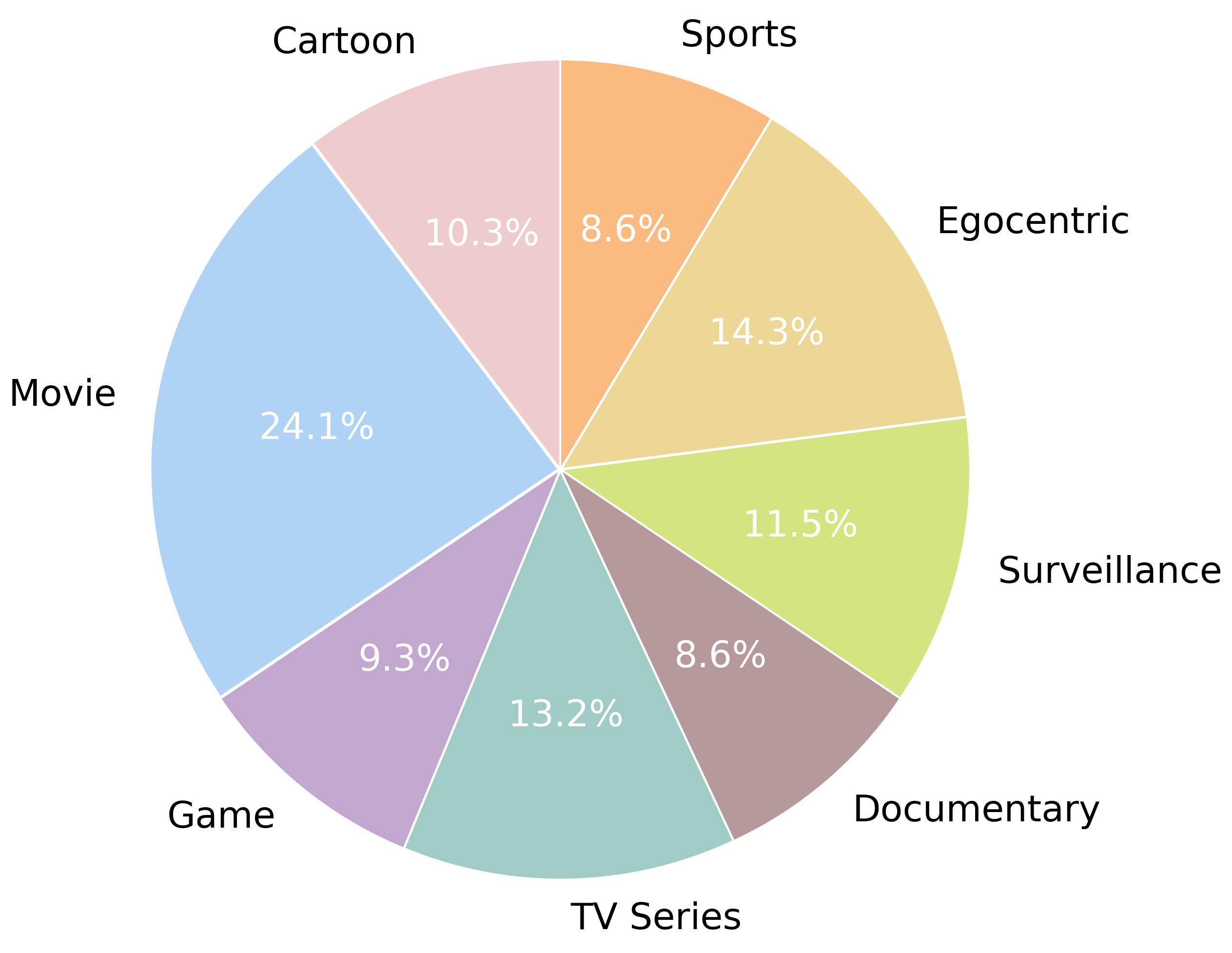}
    \caption{The distribution of video source for VICO.}
    \vspace{-5pt}
    \label{fig:source}
\end{figure}

\section{Analysis of VICO Dataset}
 \label{appendix:vico} 
To enhance long video understanding and unlock the potential of visual compression, we developed an automated long-video data production pipeline and a high-quality dataset called Visual Clue Order (VICO). In this section, we highlight its key features across several aspects.

\textbf{Diversified Video Categories.} VICO offers a comprehensive collection of videos across various genres. Initially, we source videos from CinePile, which includes movies, TV series, and cartoons. Additionally, we collect real-world videos such as egocentric videos, documentaries, games, sports, tutorials, and surveillance footage. The proportion of each video type is illustrated in Figure~\ref{fig:source}.

\textbf{Versertile video length.} VICO comprises videos of diverse lengths, ranging from 1 minute to over 9 minutes, as shown in Figure~\ref{fig:length}. Additionally, each video is annotated with QA pairs for event/action ordering and detailed captions for individual video clips. This allows MLLMs to leverage the dataset to enhance their long video comprehension capabilities.

\begin{figure}[h]
    \centering
    \includegraphics[width=0.45\textwidth]{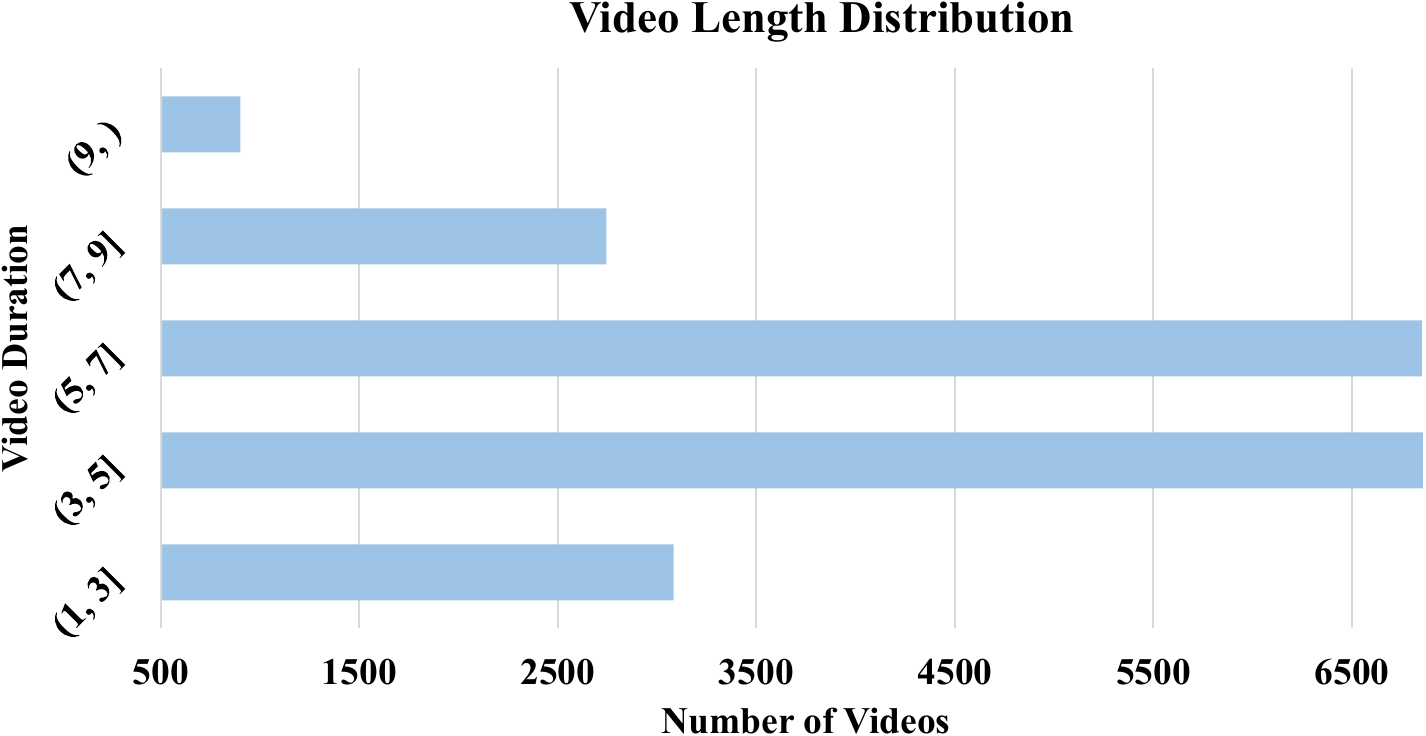}
    \caption{The distribution of video length for VICO.}
    \vspace{-5pt}
    \label{fig:length}
\end{figure}

\begin{figure*}[h]
    \centering
    % \vspace{-0.3cm}
    \includegraphics[width=0.8\textwidth]{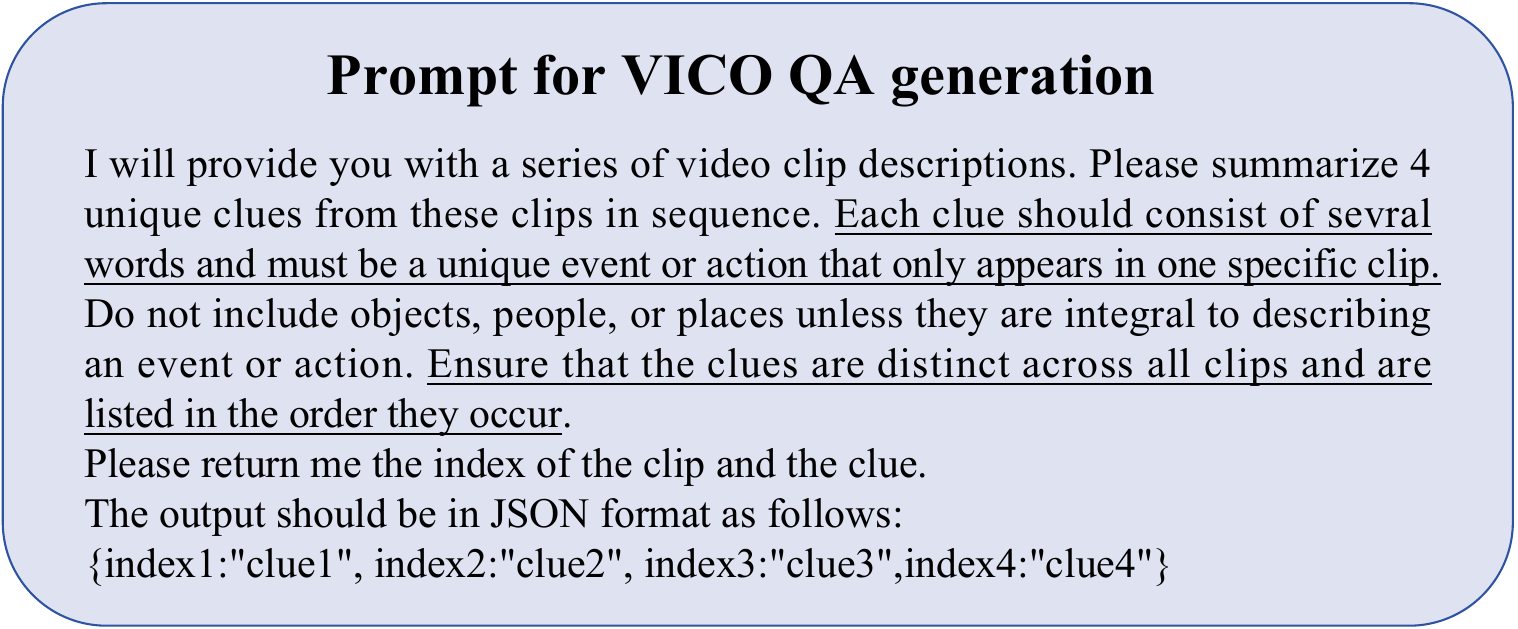}
    % \vspace{-0.5cm}
    \caption{The GPT prompt for VICO QA generation.} 
    \label{fig:prompt}
    \vspace{-0.3cm}
\end{figure*}

\begin{figure*}[h]
    \centering
    % \vspace{-0.3cm}
    \includegraphics[width=0.95\textwidth]{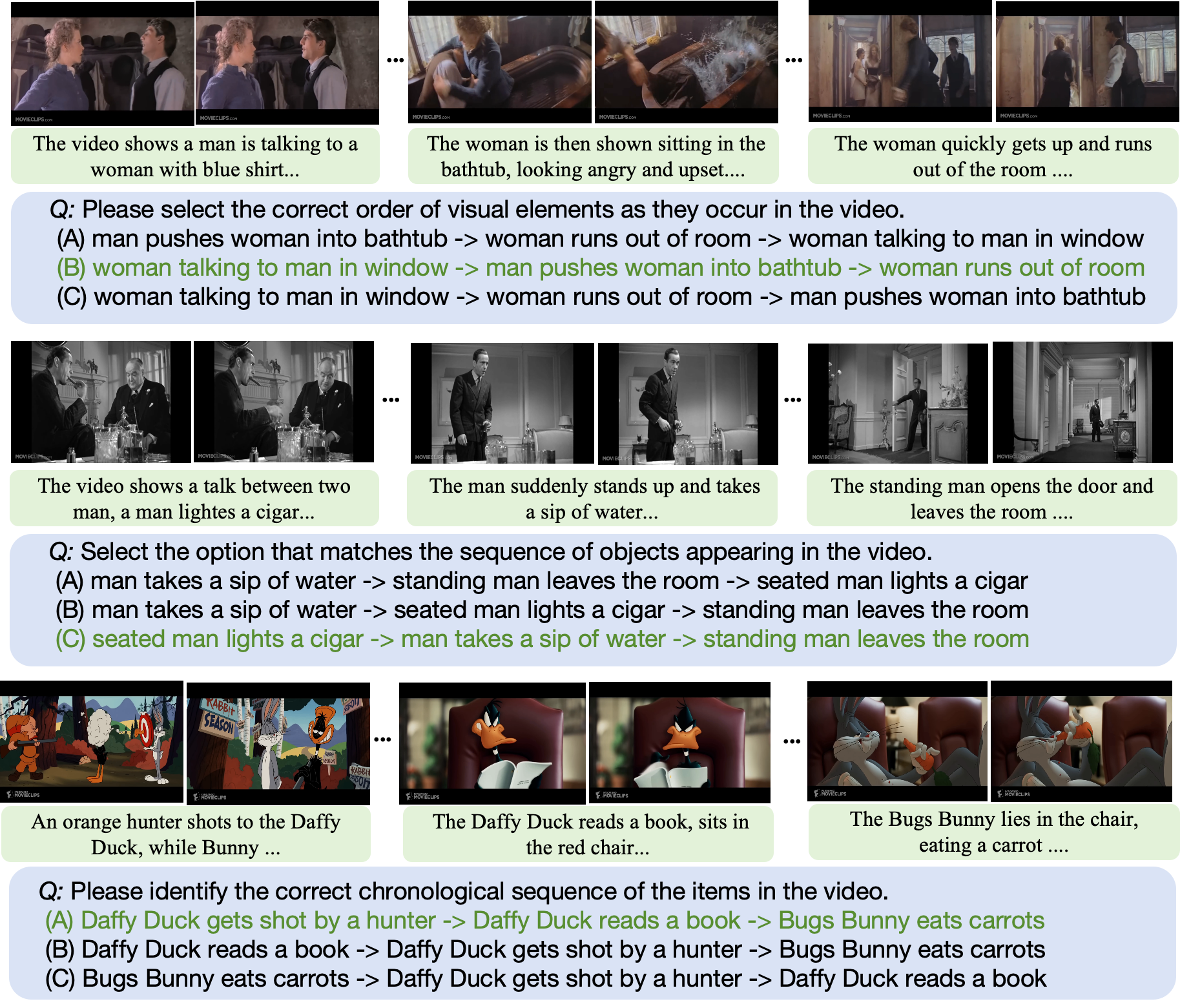}
    % \vspace{-0.5cm}
    \caption{Some visualization cases of VICO data, which includes clip cation (green) and QA pairs for event/action order (blue).} 
    \label{fig:vico_viz}
    \vspace{-0.3cm}
\end{figure*}

\begin{figure*}[h]
    \centering
    % \vspace{-0.3cm}
    \includegraphics[width=1.0\textwidth]{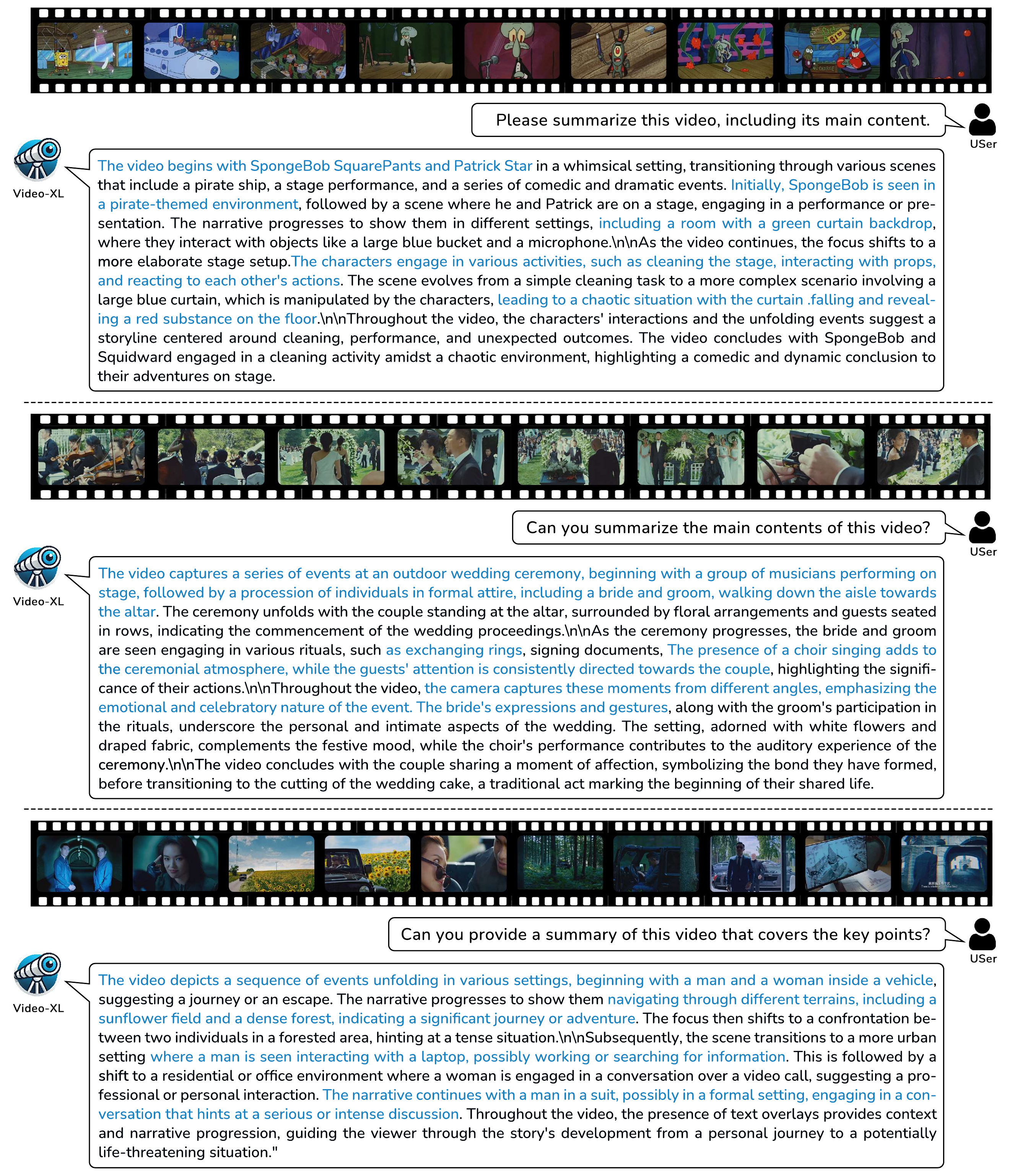}
    % \vspace{-0.5cm}
    \caption{Some visualization cases of  Video-XL on visual summarization task.} 
    \label{fig:mlvu1-3}
    \vspace{-0.3cm}
\end{figure*}
\begin{figure*}[h]
    \centering
    % \vspace{-0.3cm}
    \includegraphics[width=1.0\textwidth]{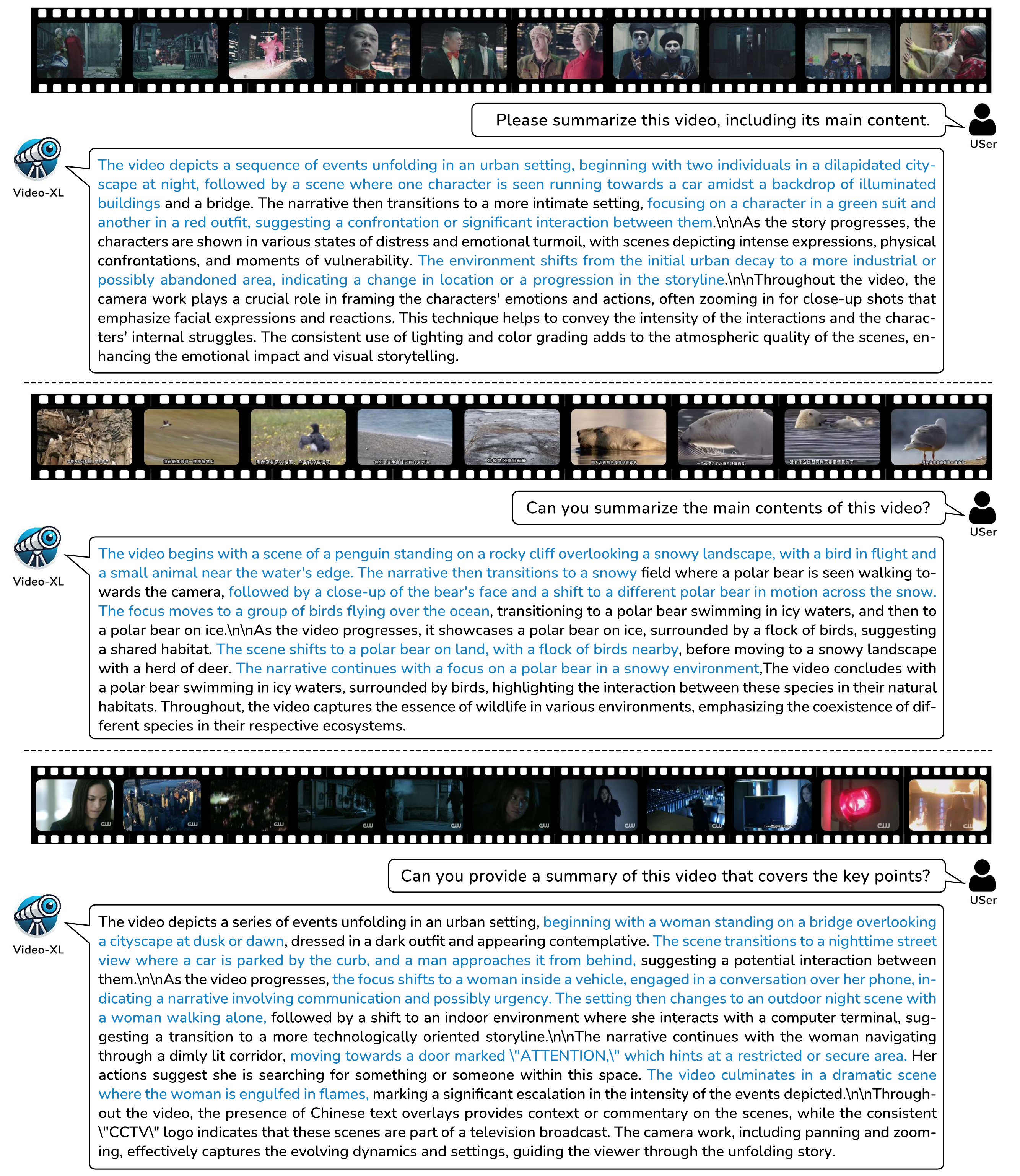}
    % \vspace{-0.5cm}
    \caption{Some visualization cases of  Video-XL on visual summarization task.} 
    \label{fig:mlvu4-6}
    \vspace{-0.3cm}
\end{figure*}

\begin{figure*}[h]
    \centering
    % \vspace{-0.3cm}
    \includegraphics[width=1.0\textwidth]{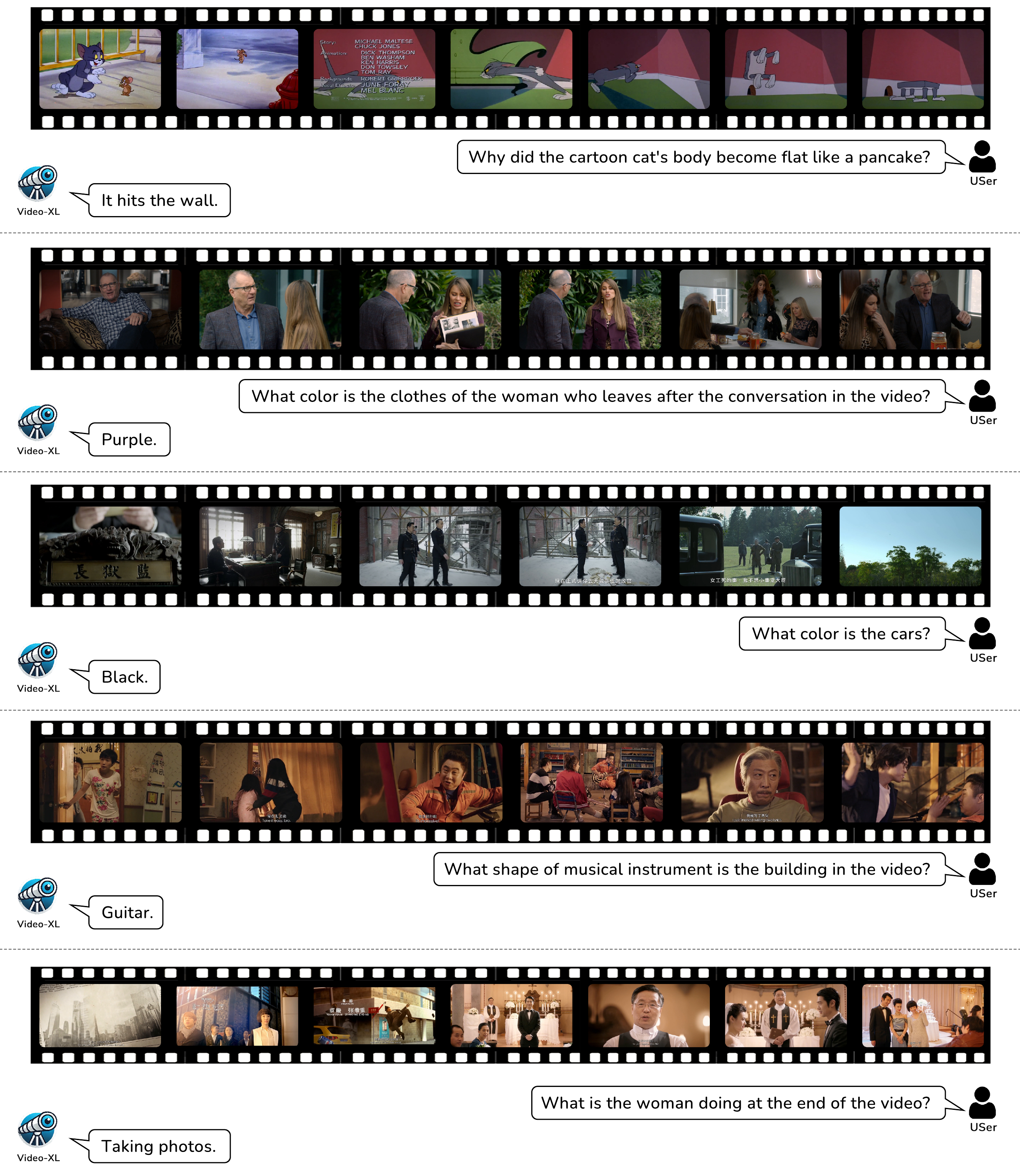}
    % \vspace{-0.5cm}
    \caption{Some visualization cases of  Video-XL on plotQA task.} 
    \label{fig:mlvu4-7}
    \vspace{-0.3cm}
\end{figure*}

\textbf{Robust Generalization.} With the assistance of VICO, Video-XL can boost its performance on general long video tasks like MLVU and Video-MME. Here, we demonstrate more experimental results, where LongVA and VideoXL are trained with different scaled VICO data. As shown in Table, both LongVA and VideoXL can benefit from the scaling up of VICO, proving that VICO strength the precise and comprehensive retrieval ability of captured information. We provide the prompt for QA generation in Figure~\ref{fig:prompt} and more visualization results of VICO in Figure~\ref{fig:vico_viz}.

\begin{table}[h]
\centering
    \centering
   
    \renewcommand{\arraystretch}{1.15}

    \begin{tabular}{>{\kern-0.5\tabcolsep}lc|cc<{\kern-0.5\tabcolsep}}
        \toprule
        \textbf{Methods}  & \textbf{Size} & \textbf{MLVU} & \textbf{Video-MME}  \\
        \midrule
         LongVA  &5k & 51.5  &59.4 \\
        LongVA  & 10k & 51.9  &60.2  \\
        LongVA  & 20k & 52.6 &61.3 \\
        \midrule
        Video-XL  & 5k & 53.9 &60.1 \\
        Video-XL  & 10k & 54.3 &60.9 \\
        Video-XL  & 20k & 54.9 &61.8 \\
        \bottomrule
    \end{tabular}
     \caption{The performance of scaling up VICO.}
    \label{tab:llm}
\end{table}

\section{Experimental Settings \& Additional Results}
\label{appendix:experiment}
We elaborate on the training and inference details of Video-XL. Since our method only modifies the workflow of LLM, the hyperparameters reported are specific to the fine-tuning stage, as shown in Table~\ref{tab:hyper}. For the inference details, we emphasize the particular context length for different benchmarks, as shown in Table~\ref{appendix:experiment}.

\begin{table}[h]
\centering
    \centering
   
    \vspace{-0.1in}
    \renewcommand{\arraystretch}{1.15}

    \begin{tabular}{>{\kern-0.5\tabcolsep}l|c<{\kern-0.5\tabcolsep}}
        \toprule
        \textbf{Hyperparameter}  & \textbf{Value}   \\
        \midrule
         Overall batch size  &8  \\
        Learning rate  & 1e-5 \\
        LR Scheduler  & Cosine decay \\
        DeepSpeed ZeRO Stage &ZeRO-2-offload \\
        Optimizer & Adam \\
        Warmup ratio & 0 \\
        Epoch & 1 \\
        Weight decay & 0 \\
        Precision & bf16 \\ 
        \bottomrule
    \end{tabular}
     \caption{Hyperparameters of Video-XL.}
    \label{tab:hyper}
\end{table}

\begin{table}[h]
\centering
    \centering
    
    \vspace{-0.1in}
    \renewcommand{\arraystretch}{1.15}

    \begin{tabular}{>{\kern-0.5\tabcolsep}l|c<{\kern-0.5\tabcolsep}}
        \toprule
        \textbf{Dataset}  & \textbf{Context Length}  \\
        \midrule
         MLVU  &256 frms  \\
        Video-MME  & 128 frms \\
        VNBench  & 1 fps \\
        LongVideoBench &256 frms \\
        VideoVista & 128 frms \\
        VideoChatGPT Bench & 16 frms\\
        MVBench & 16 frms \\
        \bottomrule
    \end{tabular}
    \caption{Experimental settings of Video-XL.}
    \label{tab:context}
\end{table}

Although Video-XL is designed for long video understanding, it also demonstrates strong proficiency in image understanding. We conduct extensive experiments on several image QA benchmarks, where Video-XL exhibits significant advantages over previous methods, as shown in Table~\ref{tab:image_results}.

\begin{table}[h]
\centering
    \centering
    
    \vspace{-0.1in}
    \renewcommand{\arraystretch}{1.15}
 \resizebox{0.98\linewidth}{!}{
    \begin{tabular}{>{\kern-0.5\tabcolsep}l|ccccc<{\kern-0.5\tabcolsep}}
        \toprule
        \textbf{Methods}  & \textbf{MME} & \textbf{MMB} & \textbf{GQA} & \textbf{POPE} &\textbf{SeedBench}  \\
        \midrule
         LLaMA-VID  &1521.4 & 65.1  &64.3 &86.0 &59.9 \\
        VoCo-LLaMA  & 1323.3 & 58.8  &57.0 &81.4 &53.7  \\
        Video-XL  & 1530.2 & 75.3 &65.1 &83.2 &59.4 \\
        \bottomrule
    \end{tabular}}
    \caption{The performance of Video-XL on mainstream image understanding benchmarks.}
    \label{tab:image_results}
\end{table}

\section{Qualitative Results}
\label{appendix:qualitative}
We present qualitative illustrations in Figures~\ref{fig:mlvu1-3}--\ref{fig:mlvu4-7}, where video samples are selected from the long video benchmark MLVU, with durations ranging from 10 to 30 minutes. To showcase the long video understanding capabilities of Video-XL, we focus on two representative tasks. The first is PlotQA, which requires the MLLM to reason about questions related to the plot of a narrative video. The second is video summarization, where the MLLM must summarize the key events in a long video. Video-XL demonstrates its ability to accurately locate relevant video segments based on a given query and provide precise answers. Additionally, it effectively captures the overall content of long videos, including summarizing main plots, describing the actions of key characters, and more.

\end{document}